\documentclass[conference]{IEEEtran}
\ifCLASSINFOpdf
\else
\fi
\usepackage{algorithm}
\usepackage{subcaption}
\usepackage{amsmath}
\DeclareMathOperator*{\argmax}{arg\,max}

\usepackage{booktabs}
\usepackage{url}
\usepackage{algpseudocode}
\algnewcommand\algorithmicinput{\textbf{Input:}}
\algnewcommand\INPUT{\item[\algorithmicinput]}
\algnewcommand\algorithmicoutput{\textbf{Output:}}
\algnewcommand\OUTPUT{\item[\algorithmicoutput]}
\usepackage{amsmath,amssymb,amsfonts}
\usepackage{graphicx}
\usepackage{textcomp}
\usepackage{xcolor}
\def\BibTeX{{\rm B\kern-.05em{\sc i\kern-.025em b}\kern-.08em
    T\kern-.1667em\lower.7ex\hbox{E}\kern-.125emX}}

\begin{document}
%
\title{Meta-AAD: Active Anomaly Detection with \\ Deep Reinforcement Learning}


\author{\IEEEauthorblockN{Daochen Zha, Kwei-Herng Lai, Mingyang Wan, Xia Hu}
\IEEEauthorblockA{Department of Computer Science and Engineering, Texas A\&M University\\
\{daochen.zha,khlai037,w1996,xiahu\}@tamu.edu}}


%


\maketitle

\begin{abstract}
High false-positive rate is a long-standing challenge for anomaly detection algorithms, especially in high-stake applications. To identify the true anomalies, in practice, analysts or domain experts will be employed to investigate the top instances one by one in a ranked list of anomalies identified by an anomaly detection system. This verification procedure generates informative labels that can be leveraged to re-rank the anomalies so as to help the analyst to discover more true anomalies given a time budget. Some re-ranking strategies have been proposed to approximate the above sequential decision process. Specifically, existing strategies have been focused on making the top instances more likely to be anomalous based on the feedback. Then they greedily select the top-1 instance for query. However, these greedy strategies could be sub-optimal since some low-ranked instances could be more helpful in the long-term. Motivated by this, in this work, we study whether modeling long-term performance can benefit active anomaly detection. This is a challenging task because it is unclear how long-term performance could be quantified. In addition, the query selection has a huge decision space, which is difficult to model. To address these challenges, we propose Active Anomaly Detection with Meta-Policy~(Meta-AAD), a novel framework that learns a meta-policy for query selection. Specifically, Meta-AAD leverages deep reinforcement learning to train the meta-policy to select the most proper instance to explicitly optimize the number of discovered anomalies throughout the querying process. Meta-AAD is easy to deploy since a trained meta-policy can be directly applied to any new datasets without further tuning. Extensive experiments on 24 benchmark datasets demonstrate that Meta-AAD significantly outperforms the state-of-the-art re-ranking strategies and the unsupervised baseline. The empirical analysis shows that the trained meta-policy is transferable and inherently achieves a balance between long-term and short-term rewards.

\end{abstract}


%
\IEEEpeerreviewmaketitle

\section{Introduction}

Anomaly detection aims to identify the data objects or behaviors that significantly deviate from the majority. Anomaly detection has essential applications in various domains, such as fraud detection, cybersecurity attack detection, and medical diagnosis~\cite{chandola2009anomaly}. Numerous anomaly detection algorithms have been proposed, but they are usually unsupervised with assumptions on the anomaly patterns~\cite{liu2008isolation,breunig2000lof}. The discrepancy between the assumptions and the real-world scenarios can lead to high false-positive rates since users may have different interests and definitions of the anomalies.

\begin{figure*}[t]
  \centering
     \begin{subfigure}[b]{0.8\textwidth}
    \includegraphics[width=1.0\textwidth]{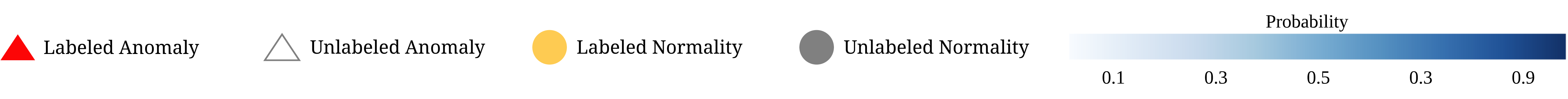}
  \end{subfigure}
  
  \begin{subfigure}[b]{0.33\textwidth}
    \includegraphics[width=0.90\textwidth]{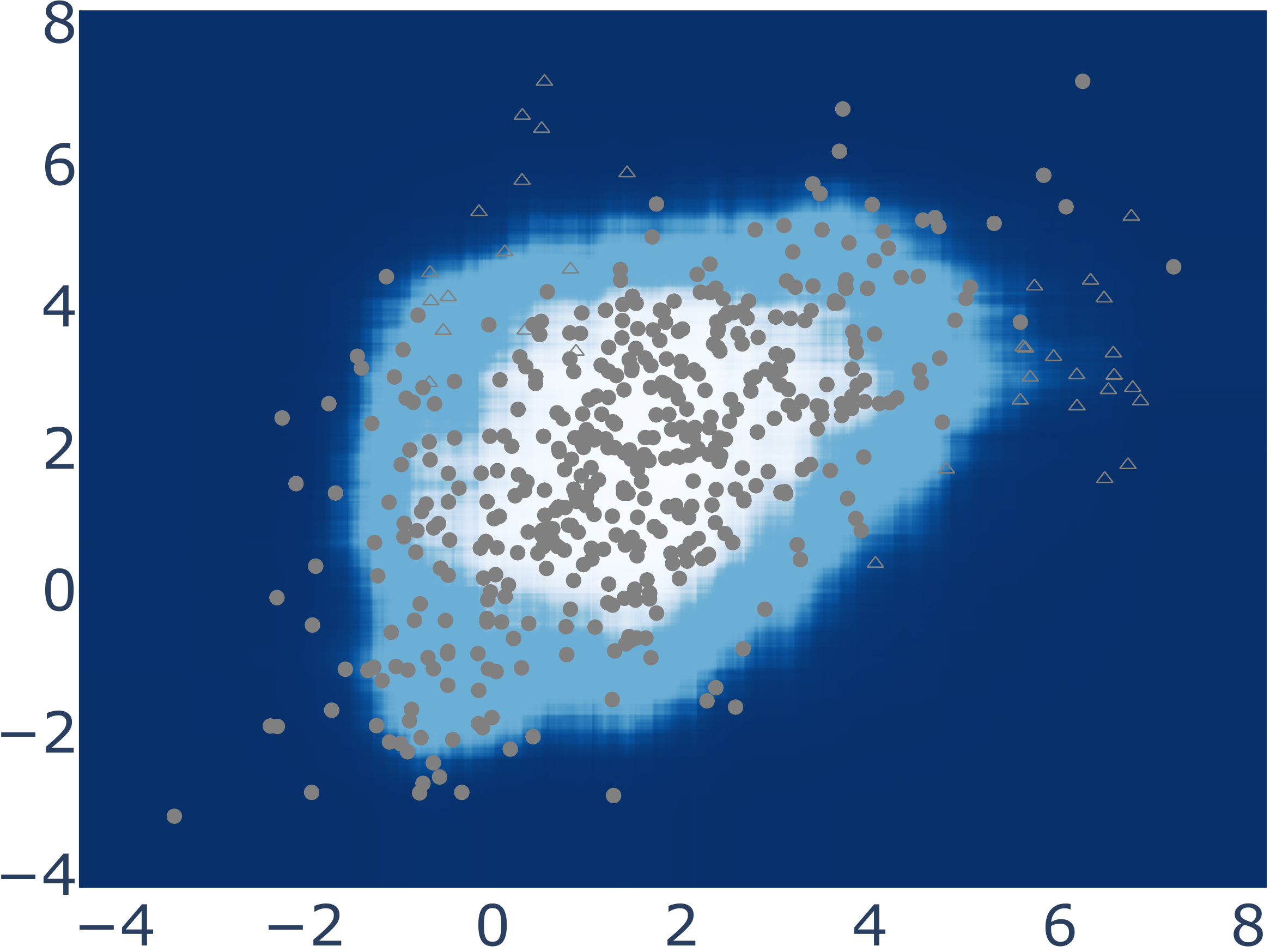}
    \caption{Initial state}
  \end{subfigure}%
  \begin{subfigure}[b]{0.33\textwidth}
    \includegraphics[width=0.90\textwidth]{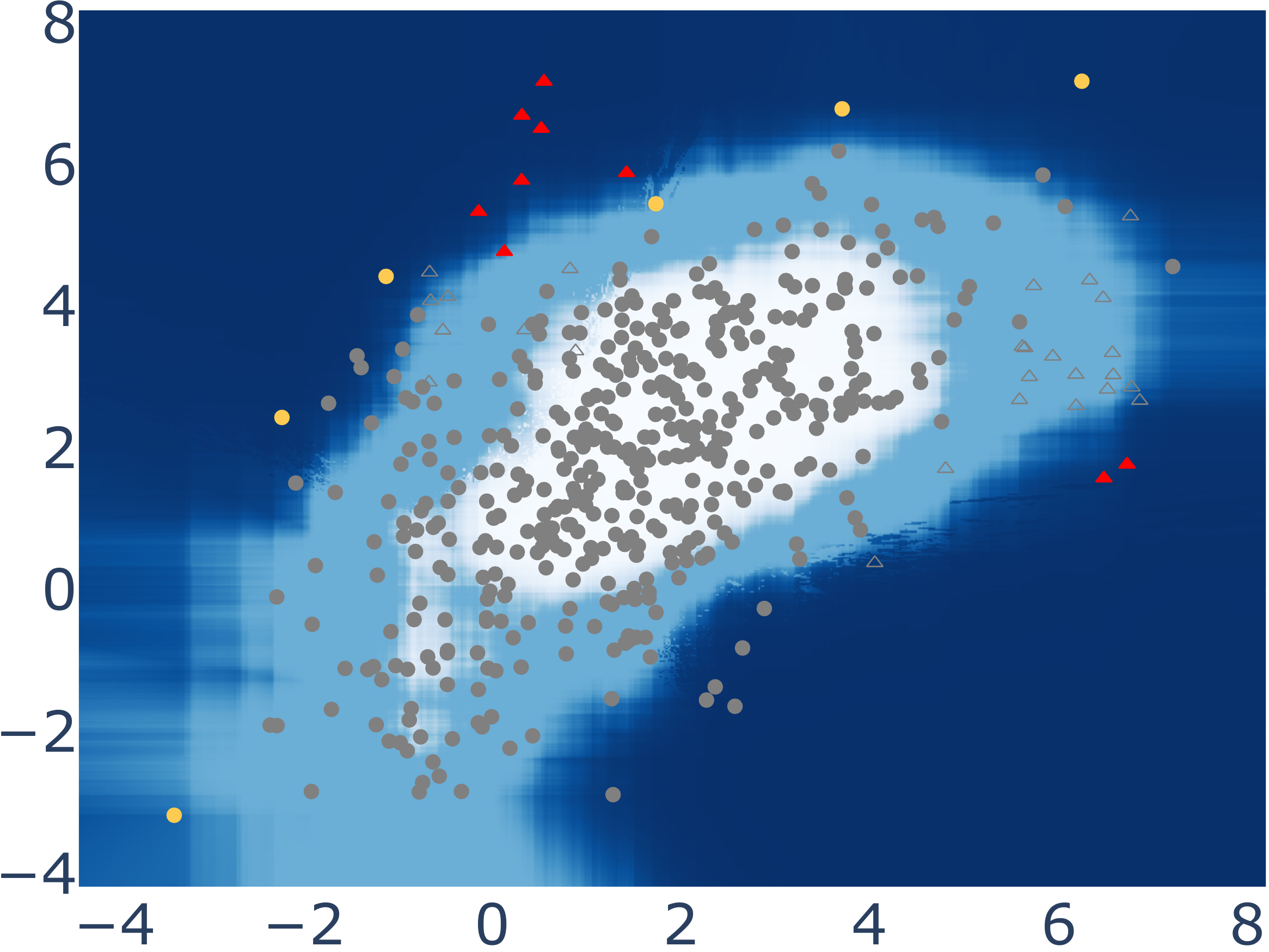}
    \caption{15 queries}
  \end{subfigure}%
  \begin{subfigure}[b]{0.33\textwidth}
    \includegraphics[width=0.90\textwidth]{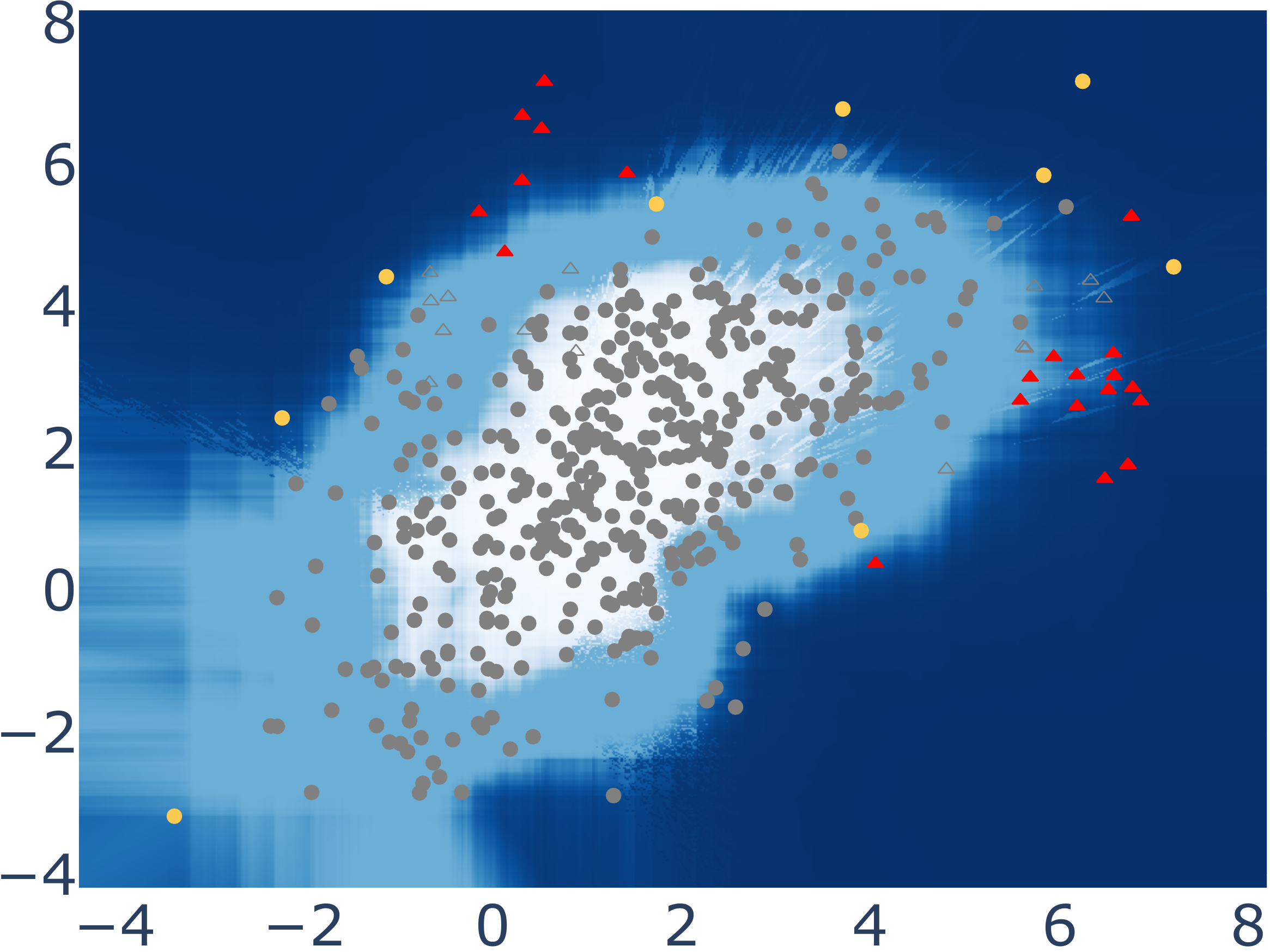}
    \caption{30 queries}
  \end{subfigure}

  \caption{Evolution of the decision of Meta-AAD on toy data. Data in blue area are more likely to be presented to the analyst. In (a), the meta-policy prefers the instances that are far away from the majority, which is similar to an unsupervised anomaly detector. In (b) and (c), with more queries, the decision pattern evolves. The probability decreases in the regions around the normal instances~(yellow). The probability increases for the regions around anomalies~(red).}
  \label{fig:toy}
\end{figure*}

In this work, we consider an alternative approach to reduce false-positive rates by involving humans in the loop. In many traditional anomaly detection scenarios, an analyst will be asked to investigate the top instances from a ranked list of anomalies to identify as many true anomalies as she can until the time budget is used up. In practice, this human feedback can be leveraged to help the analyst to identify more anomalies. We consider a scenario where the anomaly detector selects one of the instances at a time to query the analyst. Then it adjusts the decision functions by leveraging the label from the analyst. Figure~\ref{fig:toy} shows a toy example of how human feedback is leveraged to improve the detector on the toy data. We can see that human feedback can help the anomaly detector to promote the instances of interest and discourage the instances out of interest. As a result, the analyst will be presented with more true anomalies under a time budget.

Some re-ranking strategies have been proposed to approximate the above sequential decision process by greedily optimizing the immediate performance~\cite{das2016incorporating,das2017incorporating,siddiqui2018feedback,lamba2019learning}. Specifically, they adjust the anomaly scores based on the human feedback, aiming to rank anomalous instances higher. Then they greedily select the top-1 instance for the query, i.e., the one that is most likely to be anomalous. This greedy choice may benefit the immediate performance; however, it can be sub-optimal in the long-term. For example, some uncertain instances could be very helpful for correcting anomaly patterns~\cite{settles2009active}. Although these instances can be lower-ranked and harm the immediate performance, they may benefit the anomaly detector and help the analyst to discover more anomalies in future iterations. Thus, we are motivated to study whether modeling long-term performance can benefit active anomaly detection.

However, it is non-trivial to achieve this goal due to the following challenges. First, it is unclear how we can quantify the long-term performance. In the current iteration, we can only predict the intermediate outcome, i.e., whether the instances are likely to be anomalous or not, but are not clear about future benefits. Moreover, it is also difficult to balance long-term and short-term performance in different scenarios. Second, the decision space is very large since we need to examine all the instances and select one of them for the query. This makes it hard to design the selection strategy, particularly in large or high-dimensional data. Third, different datasets have various distributions of data and different sizes of decision spaces. We need a simple and transferable selection strategy that can be adopted across different datasets, which brings further challenges in designing the strategy.

To address these challenges, we propose \textbf{A}ctive \textbf{A}nomaly \textbf{D}etection with \textbf{Meta}-Policy~(Meta-AAD), which learns a meta-policy to explicitly optimize the number of discovered anomalies. Specifically, we formulate active anomaly detection as a Markov decision process and leverage deep reinforcement learning to train the meta-policy to select the most proper instance in each iteration. The meta-policy is optimized to maximize the discounted cumulative reward, which combines short-term and long-term rewards. Extensive experiments demonstrate the effectiveness of Meta-AAD, particularly in the long-term. Moreover, Meta-AAD can be easily deployed since the trained meta-policy can be directly applied to any new datasets without further tuning. The main contributions of this work are as follows.
\begin{itemize}
    \item We identify the importance of optimizing long-term performance for active anomaly detection.
    \item We propose Meta-AAD, a novel framework that leverages deep reinforcement learning to train a meta-policy to inherently optimize long-term performance.
    \item To enable the training of the meta-policy, we propose a practical solution that extracts transferable meta-features and optimizes the meta-policy on data streams.
    \item We instantiate our framework with Proximal Policy Gradients~(PPO)~\cite{schulman2017proximal}. Extensive experiments on $24$ benchmark datasets demonstrate that Meta-AAD\footnote{Code available at \url{https://github.com/daochenzha/Meta-AAD}} outperforms the state-of-the-art alternatives and the unsupervised baseline. Our empirical analysis shows that Meta-AAD can transfer across various datasets and inherently achieve a balance between long-term and short-term rewards.
\end{itemize}

\section{Preliminaries}
\label{sec:pre}
In this section, we formulate the problem of active detection with meta-policy. We then provide a background of Markov Decision Process~(MDP) and Deep Reinforcement Learning~(DRL). After that, we provide a naive approach to training the meta-policy with DRL and discuss its limitations. The main symbols used in this work are summarized in Table~\ref{tab:symbols}.

\begin{table}[]
    \centering
    \caption{Main Symbols and definitions.}
    \begin{tabular}{l|l} \toprule
    \textbf{Symbol} & \textbf{Definition} \\
    \midrule
    $n$ & The number of instances. \\
    $d$ & The feature dimension of each instance. \\
    $l$ & The dimension of transferable features. \\
    $\textbf{X} \in \mathbb{R}^{n\times d}$ & A dataset with $n$ instances and $d$ features. \\
    $\textbf{G} \in \mathbb{R}^{n\times l}$ & Transferable features with dimension $l$. \\
    $\textbf{y} \in \mathbb{R}^{n}$ & The $n$ labels of dataset, where $\textbf{y}_i \in \{-1, 1\}$.  \\
    $\hat{\textbf{y}} \in \mathbb{R}^{n}$ & The state vector, where $\hat{\textbf{y}}_i \in \{-1, 0, 1\}$.  \\
    $\textbf{c} \in \mathbb{R}^{n}$ & The anomaly scores by an unsupervised detector.  \\
    $\mathcal{S}$ & The state space in Markov Decision Process (MDP). \\
    $\mathcal{A}$ & The action space in MDP. \\
    $\mathcal{R}$ & The reward function in MDP. \\
    $\mathcal{\gamma}$ & The discount factor in MDP. \\

    \bottomrule
    \end{tabular}
    \label{tab:symbols}
\end{table}

\subsection{Problem Formulation}
\label{sec:pre1}
We consider anomaly detection problems represented by a set of instances $\textbf{X}=\{\textbf{x}_1, \textbf{x}_2,...,\textbf{x}_n\} \in \mathbb{R}^{n \times d}$, where $n$ denotes the number of instances, and $d$ denotes the feature dimension. Each instance $\textbf{x}_i$ is an $d$-dimensional vector $\{\textbf{x}_{i,1}, \textbf{x}_{i,2},...\textbf{x}_{i,d}\}$. Feature $\textbf{X}_{i,j}$ can be real-valued or categorical. Let $\textbf{y} \in \mathbb{R}^{n}$ be the ground-truths that correspond to the $n$ instances in the dataset, where $\textbf{y}_i \in \{-1,1\}$, $-1$ indicates that the instance is anomalous, and $1$ indicates that the instance is normal. Anomaly detection aims at partitioning the instances into a anomaly set $A=\{\textbf{x}_1, \textbf{x}_2,...,\textbf{x}_a\}$ and a normality set $N=\{\textbf{x}_1, \textbf{x}_2,...,\textbf{x}_b\}$, where $a$ and $b$ are the number of anomalous and normal instances, respectively. Usually, the set $A$ accounts for minority of the data, i.e., $a \ll b$.

\begin{figure*}
    \centering
    \includegraphics[width=0.97\linewidth]{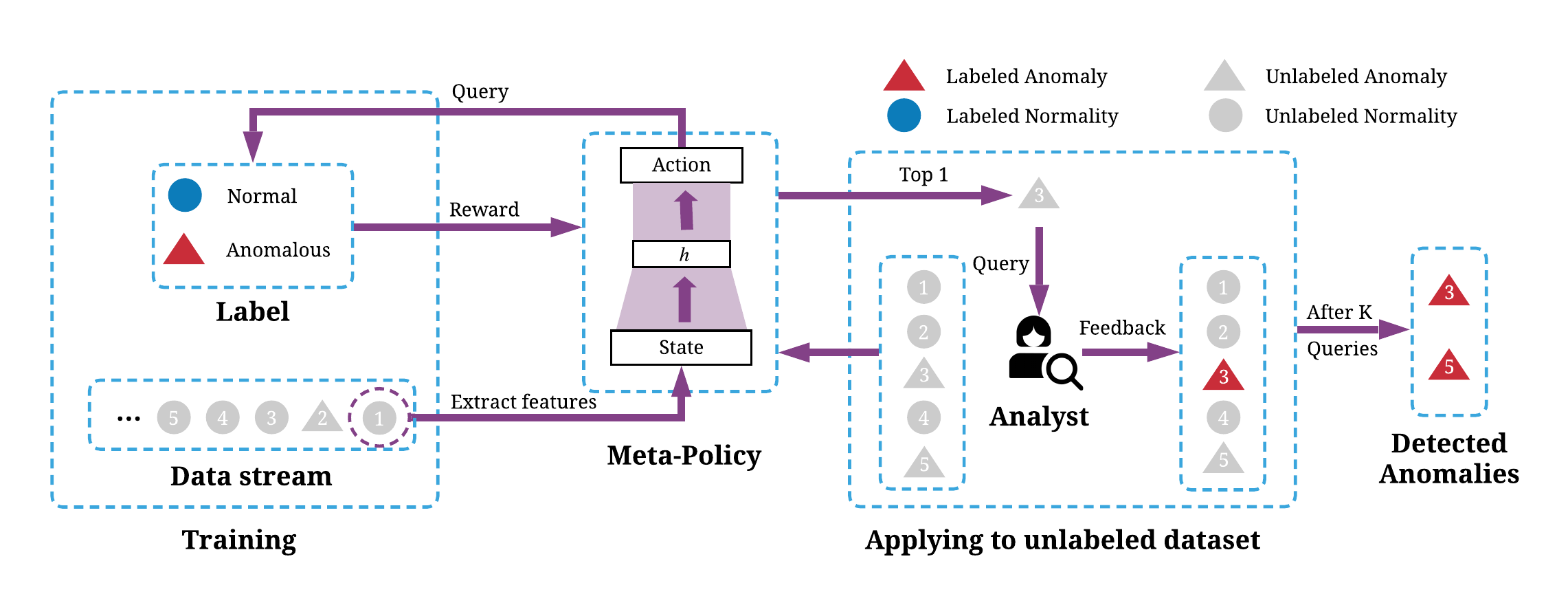}
    \caption{An overview of Meta-AAD. In training, we shuffle the data and feed them to the meta-policy in a streaming manner. The meta-policy is rewarded based on the labels. The trained meta-policy can then be directly applied to a new unlabeled dataset. In each iteration, the meta-policy chooses one of the instances and queries an analyst (human).}
    \label{fig:overview}
\end{figure*}

Conventional unsupervised anomaly detectors assign anomaly scores $\textbf{c} \in \mathbb{R}^n$ to all the instances based on $\textbf{X}$, i.e., learning a mapping $f: \textbf{X} \to \textbf{c}$, such that the lower scores indicate that the instances are more likely to be anomalous. Given the anomaly scores, we can obtain an anomaly ranking, where the anomalous instances are expected to be higher ranked than the normal instances. However, such ranking is usually not perfect since many of the higher-ranked instances may be actually normal, and some lower ranked instances could also turn out to be anomalous. Therefore, in practice, we usually require analyst (human) efforts to investigate the higher-ranked instances and decide whether they are truly anomalous or not.


Based on the notations and intuitions above, we formally describe the problem of active anomaly detection with meta-policy as follows. Given a dataset $\textbf{X}$, at each step, a meta-policy will select one of the instances $\textbf{x}_i$ for query, and a human will give a label indicating whether $\textbf{x}_i$ is truly anomalous or not. Formally, let $\hat{\textbf{y}} \in \mathbb{R}^n$ be a state vector that corresponds to the $n$ instances in the dataset. Here, $\hat{\textbf{y}}_i \in \{-1,0,1\}$, where $-1$ indicates that the instance has been selected for query and is indeed an anomaly, $1$ indicates that the instance has been selected for query but it turns out to be normal, and $0$ suggests that the instance has not been presented to the analyst yet. The state vector $\hat{\textbf{y}}$ is initialized with zeros for all the instances, i.e., no instance has been chosen for the query at the initial state. The state of the selected instance will be updated to $1$ or $-1$ at each query step based on the feedback of human. Given a budget of $T$ queries, our goal is to learn a meta-policy (trained from some other labeled datasets) to decide the instance to query at each step, i.e. a mapping $\pi: \{\textbf{X} \times \hat{\textbf{y}}\} \to \{1,2,...,n\}$, such that the number of discovered true anomalies among the chosen instances is maximized until budget $T$ is used up.

\subsection{Markov Decision Process \& Deep Reinforcement Learning}
\label{sec:pre2}
Markov Decision Process~(MDP) describes a framework for sequential decision making process. An MDP is defined as $\mathcal{M}=(\mathcal{S}, \mathcal{A}, \mathcal{P}_{T}, \mathcal{R}, \gamma)$, where $\mathcal{S}$ denotes the set of states, $\mathcal{A}$ denotes the set of actions, $\mathcal{P}_{T}: \mathcal{S} \times \mathcal{A} \times \mathcal{S} \to\mathbb{R}^+$ denotes the state transition function, $\mathcal{R}: \mathcal{S} \to \mathbb{R}$ denotes the immediate reward function, and $\gamma \in (0,1)$ is a discount factor to balance the short-term and long-term reward. At each timestep $t$, the agent takes action $a_t \in \mathcal{A}$ according to the current state $s_t \in \mathcal{S}$, and observes the next state $s_{t+1}$ as well as a reward $r_t = \mathcal{R}(s_{t+1})$. Our goal is to learn a policy $\pi: \mathcal{S} \to \mathcal{A}$ to maximize the expected discounted cumulative reward $\mathbb{E}_\pi[\sum_{t=0}^{\infty}\gamma^t r_t]$.


Deep reinforcement learning~(DRL) describes a family of algorithms for solving the MDP with deep neural networks~\cite{mnih2015human}. Contemporary DRL algorithms often learn a state value function $V(s_t)=\mathbb{E}_{a_{t},s_{t+1},...}[\sum_{l=0}^{\infty}\gamma^t \mathcal{R}(s_{t+l)}]$~\cite{schulman2015trust,schulman2017proximal} or state-action value function $Q(s_t, a_t)=\mathbb{E}_{s_{t+1},a_{t+1},...}[\sum_{l=0}^{\infty}\gamma^t \mathcal{R}(s_{t+l)}]$~\cite{mnih2015human,lillicrap2016continuous} with deep neural networks to decide the most rewarding action at each state.


\subsection{Limitations of a Naive Approach}
\label{sec:pre3}
One may come up with a naive approach to train the meta-policy with deep reinforcement learning. Specifically, the active learning process could be naturally treated as an MDP if we consider the state as the state vector and action as the queried instance, i.e., $\mathcal{S}= \{\textbf{X}\times \hat{\textbf{y}}\}$, $\mathcal{A}=\{1,2,...,n\}$. Then by appropriately defining a reward function, we can directly model the process as an MDP and train a policy to optimize performance with deep reinforcement learning algorithms.

However, this approach is infeasible because it has two limitations. First, the state and action spaces are too large. The state dimension and action dimension are $O(nd)$ and $O(n)$, respectively, since at each iteration, we can observe the information of all the $n$ instances and need to select one of the $n$ instances for query. However, the state-of-the-art deep reinforcement learning algorithms usually perform not well on large state and action spaces~\cite{dulac2015deep,sutton2018reinforcement}. In our preliminary experiments, we also observe that the above naive method fails to train an effective meta-policy. Second, even if we can train a meta-policy, it is difficult to transfer the meta-policy to another dataset since the state and action spaces are different in different datasets. The meta-policy will be of practical value only when it can be transferred. Therefore, this naive approach can not be directly applied to our problem. In the following sections, we discuss how we can address the above issues to enable stable meta-policy training.

\section{Methodology}
\label{sec:method}
In this section, we elaborate on the \textbf{A}ctive \textbf{A}nomaly \textbf{D}etection with \textbf{Meta}-Policy~(Meta-AAD). An overview of Meta-AAD is illustrated in Figure~\ref{fig:overview}. In the training stage, we extract transferable features as states~(Section~\ref{sec:method1}). We then shuffle the data and feed the data into meta-policy in a streaming manner so that the state and action spaces can be significantly reduced~(Section~\ref{sec:method2}). The meta-policy is trained with deep reinforcement learning based on some labeled datasets~(Section~\ref{sec:method3}). Finally, the trained meta-policy can be directly applied to any new unlabeled datasets for active anomaly detection without further tuning~(Section~\ref{sec:method4}).

\subsection{Extracting Transferable Meta-Features}
\label{sec:method1}

In this subsection, we aim to extract transferable meta-features that can be used across different datasets, i.e., we aim at defining a mapping $g: \{\textbf{X} \times \hat{\textbf{y}}\} \to \textbf{G} \in \mathbb{R}^{n \times l}$, where $l$ is the dimension of extracted features, such that $\textbf{G}$ is less dependent on the dataset.

Intuitively, there are three types of information that are critical for deciding which instance to query. The first is anomaly scores outputted by the anomaly detector. Anomaly scores can provide information about which instances are far away from the majorities to help the meta-policy to discover more anomalous instances. Second, the labeled anomalous instances are helpful. With several queries, we may be able to identify some anomalous instances. Properly promoting the instances that are similar to these known anomalous instances will improve the performance. Third, labeled normal instances are also useful. Similarly, discouraging the instances that are similar to the known normal instancs may decrease the false positives. Based on the intuitions above, we empirically extract some features as follows, with a total of $6$ features.

\begin{itemize}
    \item \textbf{Detector features:} The anomaly scores $\textbf{c}$ outputted by unsupervised anomaly detectors. Any off-the-shelf anomaly detection algorithms can serve as detectors.
    \item \textbf{Anomaly features:} The features indicating the relatedness to the labeled anomalous instances. In this work, we extract three features for this purpose. We standardize the original features $\textbf{X}$ and calculate the minimum and the mean Euler distances to the labeled anomalous instances. In addition, we introduce a binary feature indicating whether there exists an anomalous instance in the $k$-nearest neighbors or not.
    \item \textbf{Normality features:} Similarly, we use the minimum and the mean Euler distances to the labeled normal instances as the normality features.
\end{itemize}
Note that our framework allows flexible choices of features. For example, we may be able to improve the performance by using an ensemble of unsupervised anomalous detectors or more fine-grained anomaly and normality features. To make our contribution focused, we adopt these simple features in all our experiments, which lead to reasonable performance based on our empirical results. How we can better model the transferable information will be an interesting future work to enhance the meta-policy.

By mapping the original features to the above transferable features, we will have the same feature dimension in different datasets, i.e., $l$ is the same. However, the new features are not ready to be used for training since different datasets have a different number of instances $n$. We will address this remaining issue in the next subsection.

\subsection{Learning from Data Streams}
\label{sec:method2}
The transferable features $\textbf{G}\in \mathbb{R}^{n \times g}$ obtained in the previous section and the action space $\mathcal{A}=\{1,2,...,n\}$ are still too large for a learning algorithm. Moreover, the size of the spaces is proportional to the size of the dataset, which makes the meta-policy impossible to transfer.

To enable the training of transferable meta-policy, we propose to instead operate on data streams. Specifically, given the transferable features of a training data $\textbf{G}^{train}$ and its corresponding labels $\textbf{y}^{train}$. In each episode, we randomly shuffle $\textbf{G}^{train}$ and $\textbf{y}^{train}$ to obtain a perturbation, denoted as $\textbf{G}^{train'}$ and $\textbf{y}^{train'}$. Instead of giving all the data to the meta-policy, we feed the meta-policy with one instance at a time. In the streaming setting, the state, action and reward of the Markov Decision Process~(MDP) are defined as follows.

\begin{itemize}
    \item \textbf{State $\mathcal{S}$:} The transferable features of the current observed instance $\textbf{G}^{train'}_i \in \mathbb{R}^{l}$, where $i$ is the instance index.
    \item \textbf{Action $\mathcal{A}$:} Actions can be $1$ or $0$, where $1$ suggests that the current instance should be selected, while $0$ suggests that current instance should be ignored.
    \item \textbf{Reward $\mathcal{R}$:} If the meta-policy queries an instance, we give a positive reward of $1$ if the instance is indeed anomalous, and a small negative reward of $-0.1$ if the instance is normal. We give $0$ reward if the meta-policy ignores an instance. The reward function is critical to describe the desired behaviors. We will empirically study the impact of different reward choices in the experiments~(see the bottom of Figure~\ref{fig:ablation}).
\end{itemize}
The above MDP describes an active learning procedure in a streaming setting. Intuitively, the meta-policy is encouraged to take action $1$ if the queried instance is anomalous and take action $0$ if the queried instance is normal. In this sense, the meta-policy will be taught to discover more anomalies under a budget. We note that the meta-policy trained in a streaming setting could be sub-optimal when applied to the batch setting since the two MDPs have different objectives. Nonetheless, we find in practice that this concern is greatly outweighed by the benefits that the streaming setting brings. It significantly reduces the state and action spaces to make the training of transferable meta-policy feasible.

\begin{algorithm}[t]
\caption{Training meta-policy with PPO}
\label{alg:1}

\begin{algorithmic}[1]
 \INPUT A set of features $\{\textbf{X}^i\}_{i=1}^{N}$ and the corresponding labels $\{\textbf{y}^i\}_{i=1}^{N}$, rollout steps $T$
 \OUTPUT The trained meta-policy
\State Initialize meta-policy $\pi_{\theta}$, $\theta_{old} \leftarrow \theta$
\For{iteration = $1$, $2$, ... until convergence}
    \If {iteration = 1 or episode is over}
        \State Randomly sample $\{\textbf{X}', \textbf{y}'\}$ from $\{\textbf{X}^i\}_{i=1}^{N}$, $\{\textbf{y}^i\}_{i=1}^{N}$
    \EndIf
    \State Run $\pi_{\theta_{old}}$ with $\{\textbf{X}', \textbf{y}'\}$ based on the MDP defined in Section~\ref{sec:method1} for $T$ timesteps.
    \State Compute advantages $\hat{A}_1, ..., \hat{A}_t$ based on Equation~(\ref{eqn:advantage})
    \State Update $\theta$ based on Equation~(\ref{eqn:update})
    \State $\theta_{old} \leftarrow \theta$
\EndFor \\
\Return $\pi_\theta$
\end{algorithmic}
\end{algorithm}

\subsection{Training Meta-Policy with Deep Reinforcement Learning}
\label{sec:method3}
Given the MDP defined in Section~\ref{sec:method2}, we can train the meta-policy with any deep reinforcement learning~(DRL) algorithms. In this work, we instantiate our framework with Proximal Policy Optimization~(PPO)~\cite{schulman2017proximal}. We note that there are more advanced algorithms, such as~\cite{haarnoja2018soft}, which we will explore in the future.

The meta-policy is described as a parametric policy $\pi_\theta(a|s)$, where $s$ is an $l$ dimensional feature, $a\in\{0,1\}$, $\sum_{a \in \{0,1\}} \pi (a|s) = 1$, and $\theta$ is the parameters of the network. Our goal is to maximize the discounted cumulative reward $\mathbb{E}_\pi[\sum_{t=0}^{\infty}\gamma^t r_t]$. PPO is an actor-critic algorithm, where the critic approximates the state values and the actor is the policy. Specifically, the critic of PPO trains a deep neural network to approximate $V(s)$ through interacting with the environment. Then a generalized advantage estimator~\cite{schulman2015high} is used:
\begin{equation}
    \hat{A}_t = \delta_t + \sum_{t'=1}^{T-1+1}(\gamma \lambda)^{t'} \delta_{t+t'},
    \label{eqn:advantage}
\end{equation}
where $\delta_t = r_t + \gamma V(s_{t+1}) - V(s_t)$, $T$ is the total timesteps in an episode, $\gamma$ is the discount factor, and $\lambda$ is a hyper-parameter to control the bias-variance trade-off. Intuitively, advantage values measure how much an action is better than the other actions. Based on the estimated advantages, the actor is updated by a clipped surrogate objective:
\begin{equation}
    L_t^{CLIP} (\theta) = \hat{\mathbb{E}}_t[\min(r_t(\theta) \hat{A}_t, clip(r_t(\theta), 1-\epsilon, 1+\epsilon) \hat{A}_t)],
\end{equation}
where $r_t(\theta) = \frac{\pi_\theta(a_t|s_t)}{\pi_{\theta_{old}}(a_t|s_t)}$, $\pi_{\theta_{old}}$ is the policy before the update, $clip(r_t(\theta), 1-\epsilon, 1+\epsilon)$ will clip $r_t(\theta)$ into range $[ 1-\epsilon, 1+\epsilon]$, and $\epsilon$ is a hyper-parameter to control the clip range. The clipping objective makes sure that the new policy will not deviate too much from the old policy, which enables stable policy improvement. In training, we use a combined loss to simultaneously update the value loss:
\begin{equation}
    L_t(\theta) = \hat{\mathbb{E}}_t [L_t^{CLIP}(\theta)-c_1 L_t^{VF}(\theta)+ c_2 \cdot entropy(\pi_\theta(\cdot|s_t))],
    \label{eqn:update}
\end{equation}
where $L_t^{VF}(\theta)$ is a squared-error loss $(V_\theta(s_t)-V_t^{target})^2$, $V_t^{target}$ is estimated based on the collected data, $entropy(\cdot)$ is a term to encourage exploration, $c_1$ and $c_2$ are hyper-parameters. The expectation in Equation~(\ref{eqn:update}) can be approximated by sampling data from the environment.

\begin{algorithm}[t]
\caption{Application of trained meta-policy}
\label{alg:2}
\begin{algorithmic}[1]
\INPUT Unlabeled dataset $\textbf{X} \in \mathbb{R}^{n \times d}$, trained meta-policy $\pi_\theta$
\OUTPUT The detected anomalies
\State Initialize state vector $\hat{\textbf{y}} = \{0\}_{i=1}^{n}$, anomalous list $\textbf{A} = \{\}$
\For{iteration = $1$, $2$, ... until budget is used up}
    \State Obtain transferable features $\textbf{G} \in \mathbb{R}^{n \times l}$ from $\{\textbf{X}, \hat{\textbf{y}}\}$
    \State Compute $\pi(a=1|s)$ based on $\textbf{G}$ as $\textbf{p} \in \mathbb{R}^{n}$
    \State Query the instance with the highest probability
    \If {the instance is anomalous}
        \State Put the instance into $\textbf{A}$
    \EndIf
    \State Update $\hat{\textbf{y}}$ based on human feedback
\EndFor\\
\Return $\textbf{A}$
\end{algorithmic}
\end{algorithm}

The training procedure of the meta-policy is summarized in Algorithm~\ref{alg:1}. We assume the availability of several labeled datasets. In each episode, we randomly choose a dataset, shuffle the instances and traverse the dataset from the beginning in a streaming manner.

\subsection{Application of Meta-Policy}
\label{sec:method4}
Once the meta-policy is trained, we can directly apply it to any new unlabeled datasets without further tuning. However, we note that there are some major differences between the application and the training. First, instead of feeding one feature to the meta-policy at a time, we give all the features to the meta-policy to compute the probabilities for all the instances. Specifically, when applying the meta-policy to an unlabeled dataset $\textbf{X}\in \mathbb{R}^{n \times d}$, we first extract the transferable features $\textbf{G} \in \mathbb{R}^{n \times l}$ according to $\textbf{X}$ and the current state vector $\hat{\textbf{y}} \in \mathbb{R}^{n}$. Then we compute $\pi_\theta(a=1|\textbf{G}_i), \forall{i \in \{1,2,...,n\}}$ and obtain the probabilities $\textbf{p} \in \mathbb{R}^{n}$. Then we choose the instance with highest probability for query, i.e., $\argmax_{i}\textbf{p}_i$. Intuitively, the instance that is very likely to be selected in the streaming setting is also very likely to be chosen in this batch setting. The above procedure is summarized in Algorithm~\ref{alg:2}.

Note that $\pi_\theta(a=1|\textbf{G}_i)$ is fundamentally different from the adjusted anomaly scores. In previous methods~\cite{das2017incorporating,das2016incorporating,siddiqui2018feedback,lamba2019learning}, the anomalous scores are adjusted to promote the anomalous instances to the top. The main goal of the adjustment is to make the top-1 instance more likely to be anomalous so as to maximize the immediate performance. Whereas, the probability of the meta-policy plays a significantly different role. The probability is learned with the objective of maximizing the discounted cumulative reward, which is a combination of immediate and long-term rewards. That is, the long-term performance is inherently incorporated into the probabilities and the top-1 selection strategy.

\section{Experiments}
\label{sec:exp}
In this section, we conduct extensive experiments to evaluate Meta-AAD. We mainly focus on the following research questions.

\begin{itemize}
    \item \textbf{RQ1:} How does the meta-policy select the query and how will the decision of the meta-policy evolve in different stages~(Section~\ref{sec:exp2})?
    \item \textbf{RQ2:} How does Meta-AAD compare with the state-of-the-art alternatives and unsupervised baseline~(Section~\ref{sec:exp3})?
    \item \textbf{RQ3:} How will Meta-AAD perform if using different features, the number of labeled datasets and reward functions~(Section~\ref{sec:exp4})?
    \item \textbf{RQ4:} How many computational resources are needed to train a meta-policy~(Section~\ref{sec:exp5})?
    \item \textbf{RQ5:} How does Meta-Policy balance long-term and short-term reward~(Section~\ref{sec:exp5})?
\end{itemize}

\subsection{Experimental Settings}
\label{sec:exp1}
\textbf{Datasets and evaluation metric.} To demonstrate the generality of Meta-AAD, we select $24$ datasets with various sizes, feature dimensions and anomaly ratios from ODDS\footnote{\url{http://odds.cs.stonybrook.edu/}}. Table~\ref{tab:dataset} summarizes the statistics of the datasets. We also use a toy dataset from~\cite{das2017incorporating} for better visualization. For the evaluation metric, we use anomaly discovery curve~\cite{ding2019interactive}, which plots the number of discovered anomalies with respect to the number of queries. A perfect result is a line with a slope $1$, i.e., all the queries are anomalous. The worst case is a line with a slope $0$, i.e., all the queries are normal. Following~\cite{siddiqui2018feedback}, we set the maximum budget to be $100$ for all the datasets.

\textbf{Baselines.} We compare Meta-AAD with the state-of-the-art methods as well as an unsupervised baseline as follows.
\begin{itemize}
   \item \textbf{AAD.} Active Anomaly Detection~\cite{das2017incorporating} is a state-of-the-art method based on node re-weighting.
    \item \textbf{FIF.} Feedback-Guided Isolation Forest~\cite{siddiqui2018feedback} is a recently proposed active anomaly detector via online optimization.
    \item \textbf{SSDO.} Semi-Supervised Detection of Outliers~\cite{vercruyssen2018semi} is a recent semi-supervised point-wise anomaly detector. We are interested in studying how semi-supervised methods will perform in the active learning setting since they are also designed to leverage label information.
    \item \textbf{Unsupervised.} We also include Isolation Forest~(IF)~\cite{liu2008isolation} as an unsupervised baseline.
\end{itemize}
While our Meta-AAD can be generally applied to any unsupervised anomaly detectors or an ensemble of detectors, for a fair comparison, we follow the previous work~\cite{das2017incorporating,siddiqui2018feedback} and use Isolation Forest~(IF)~\cite{liu2008isolation} with the same hyper-parameters as in~\cite{das2017incorporating,siddiqui2018feedback}. For SSDO and the unsupervised baseline, we select the top-1 anomalous instance in each iteration.

\begin{table}[]
    \centering
    \caption{Statistics of the datasets.}
    \begin{tabular}{l|cccc} \toprule
    \textbf{Dataset} & \textbf{Points} & \textbf{Dim.} & \textbf{Anomalies} & \textbf{Anomaly\%} \\ \midrule
    Annthyroid & 7200 & 6 & 534 & 7.4 \\
    Arrhythmia & 452 & 274 & 66 & 15.0 \\
    Breastw & 683 & 9 & 239 & 35.0 \\
    Cardio & 1831 & 21 & 176 & 9.6 \\
    Glass & 214 & 9 & 9 & 4.2 \\
    Ionosphere & 351 & 33 & 126 & 36.0 \\
    Letter & 1600 & 32 & 100 & 6.3 \\
    Lympho & 148 & 18 & 6 & 4.1 \\
    Mammography & 11183 & 6 & 260 & 2.3 \\
    Mnist & 7603 & 100 & 700 & 9.2 \\
    Musk & 3062 & 166 & 97 & 3.2 \\
    Optdigits & 5216 & 64 & 150 & 3.0 \\
    Pendigits & 6870 & 16 & 156 & 2.3 \\
    Pima & 768 & 8 & 268 & 35 \\
    Satellite & 6435 & 36 & 2036 & 32.0 \\
    Satimage-2 & 5803 & 36 & 71 & 1.2 \\
    Shuttle & 49097 & 9 & 3511 & 7.0 \\
    Speech & 3686 & 400 & 61 & 1.7 \\
    Thyroid & 3772 & 6 & 93 & 2.5 \\
    Vertebral & 240 & 6 & 30 & 12.5 \\
    Vowels & 1456 & 12 & 50 & 3.4 \\
    Wbc & 278 & 30 & 21 & 5.6 \\
    Wine & 129 & 13 & 10 & 7.7 \\
    Yeast & 1364 & 8 & 64 & 4.7 \\
    \bottomrule
    \end{tabular}
    \label{tab:dataset}
\end{table}

\begin{figure*}
  \centering
  \begin{subfigure}[b]{0.50\textwidth}
    \includegraphics[width=\textwidth]{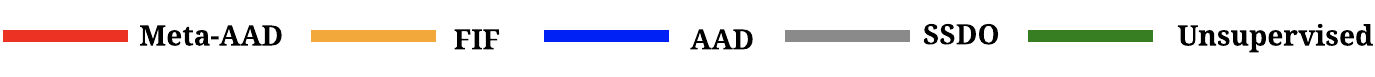}
  \end{subfigure}

  \begin{subfigure}[b]{0.166\textwidth}
    \includegraphics[width=\textwidth]{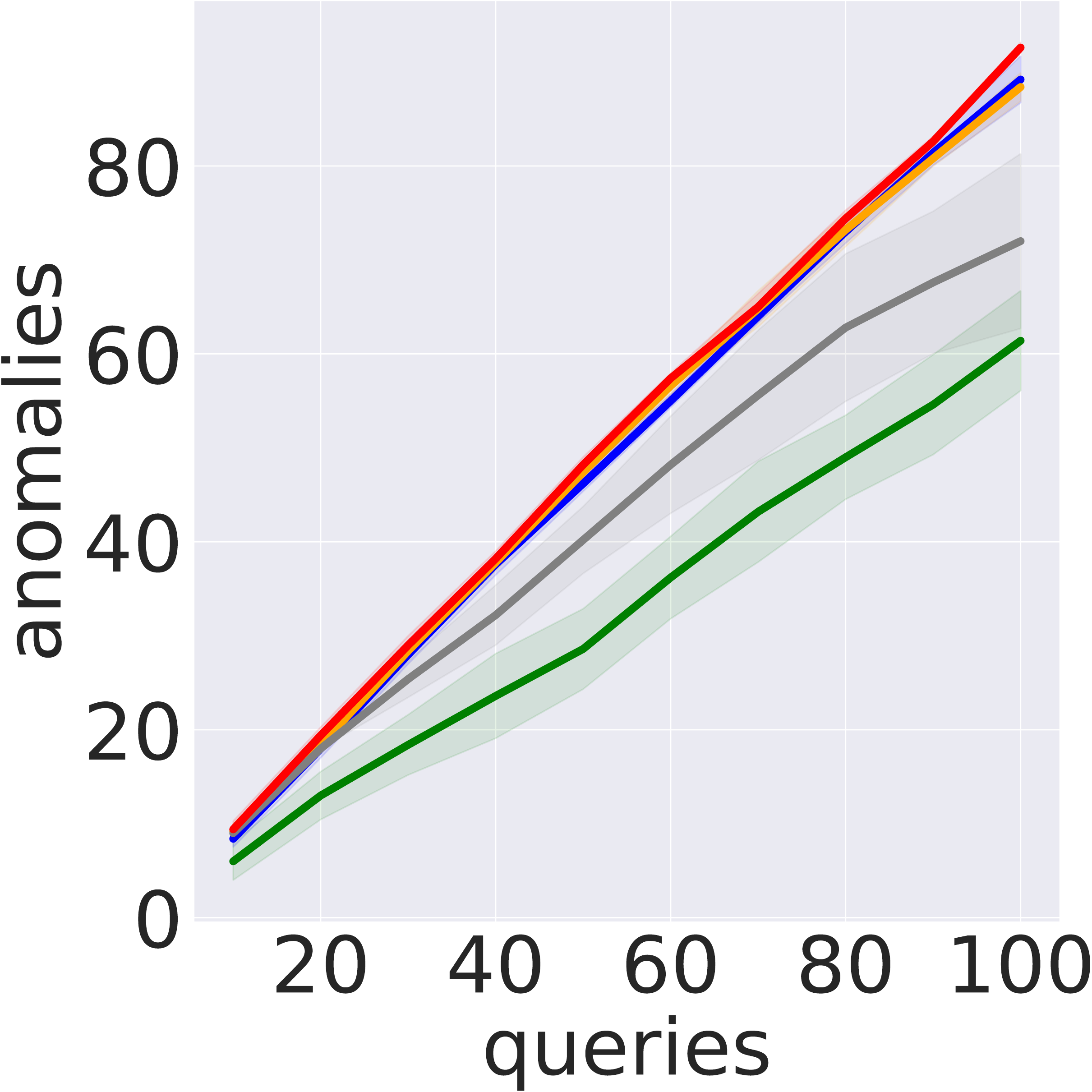}
    \caption{Annthyroid}
  \end{subfigure}%
  \begin{subfigure}[b]{0.166\textwidth}
    \includegraphics[width=\textwidth]{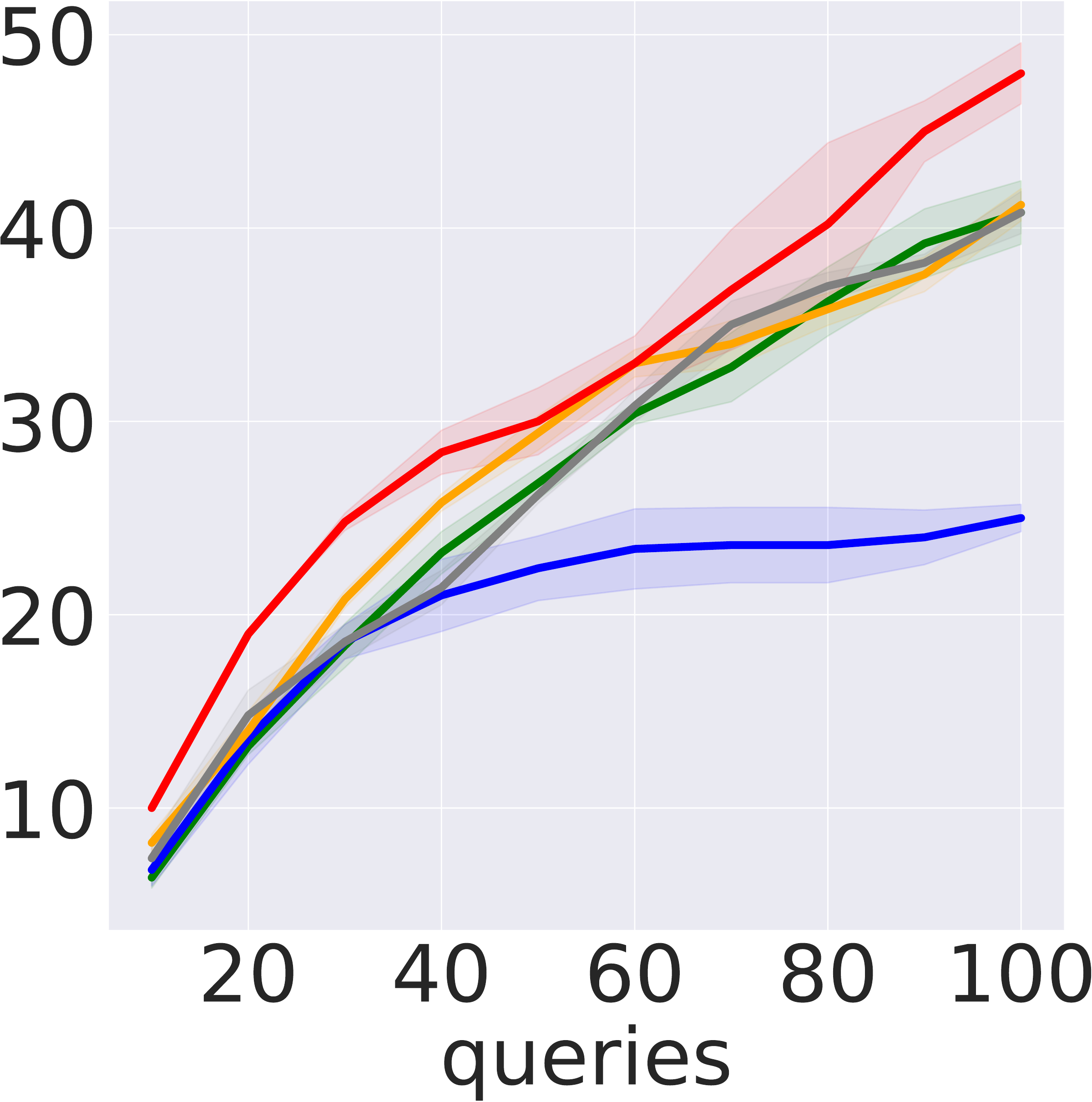}
    \caption{Arrhythmia}
  \end{subfigure}%
  \begin{subfigure}[b]{0.166\textwidth}
    \includegraphics[width=\textwidth]{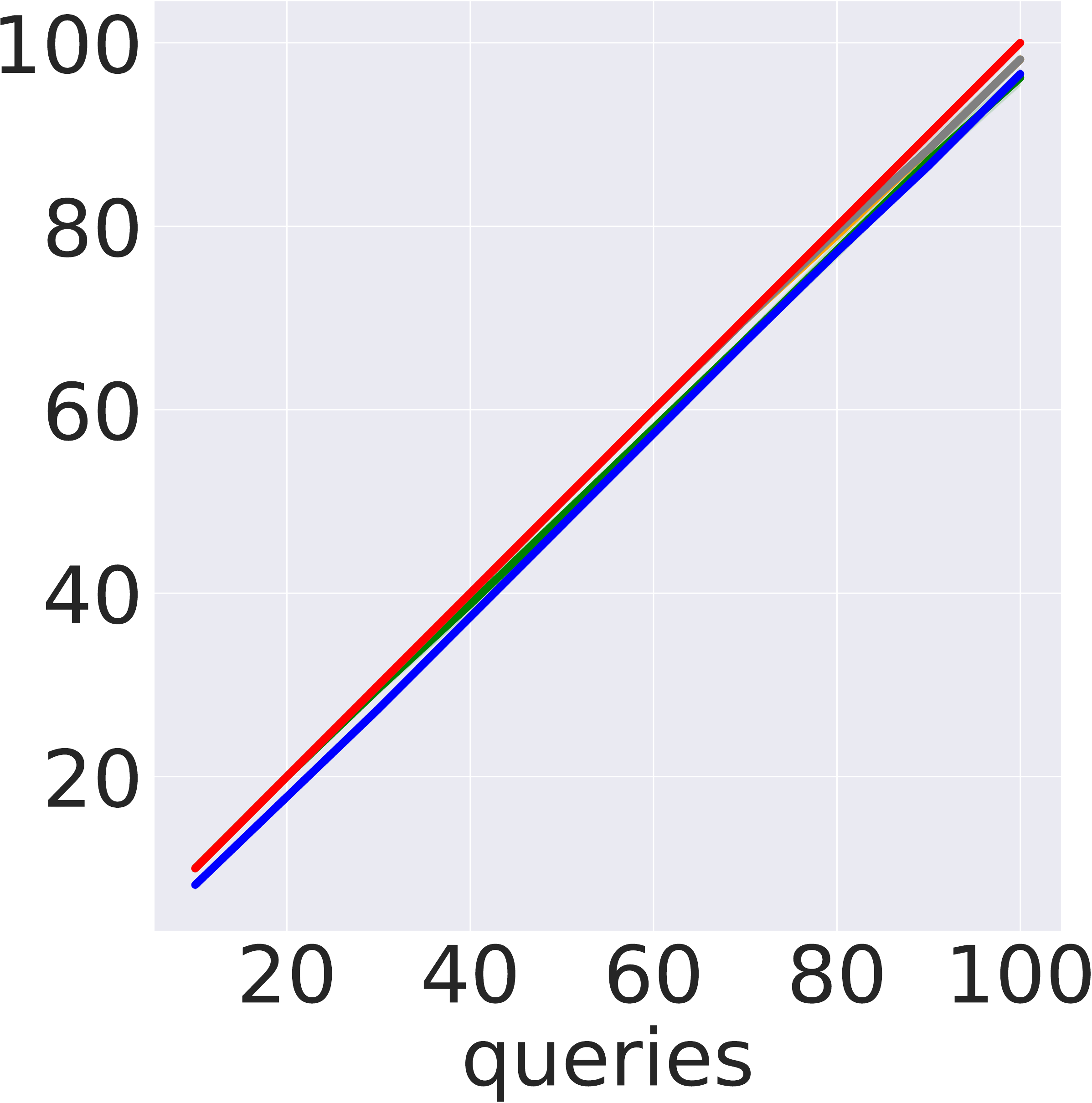}
    \caption{Breastw}
  \end{subfigure}%
  \begin{subfigure}[b]{0.166\textwidth}
    \includegraphics[width=\textwidth]{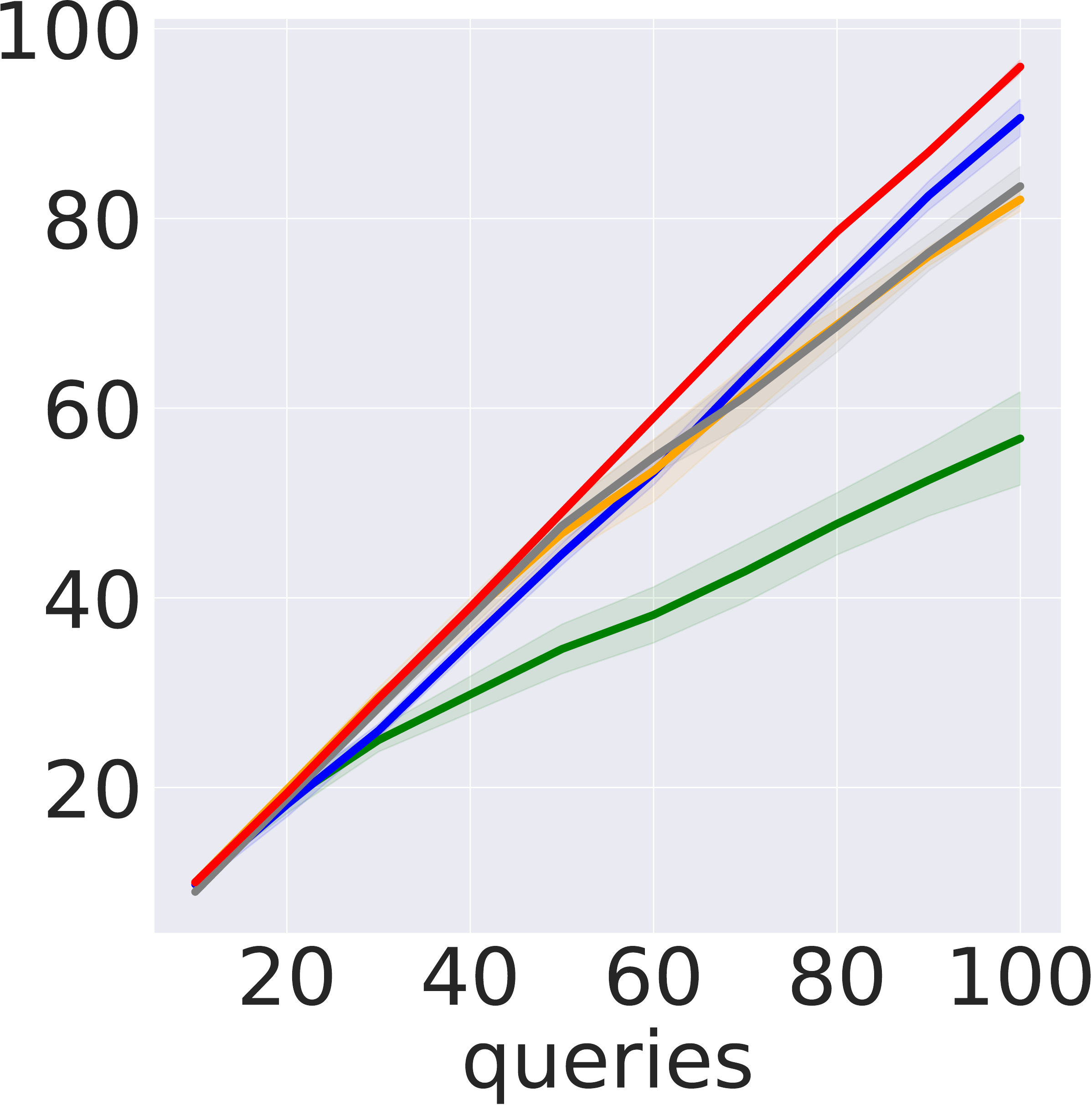}
    \caption{Cardio}
  \end{subfigure}%
  \begin{subfigure}[b]{0.166\textwidth}
    \includegraphics[width=\textwidth]{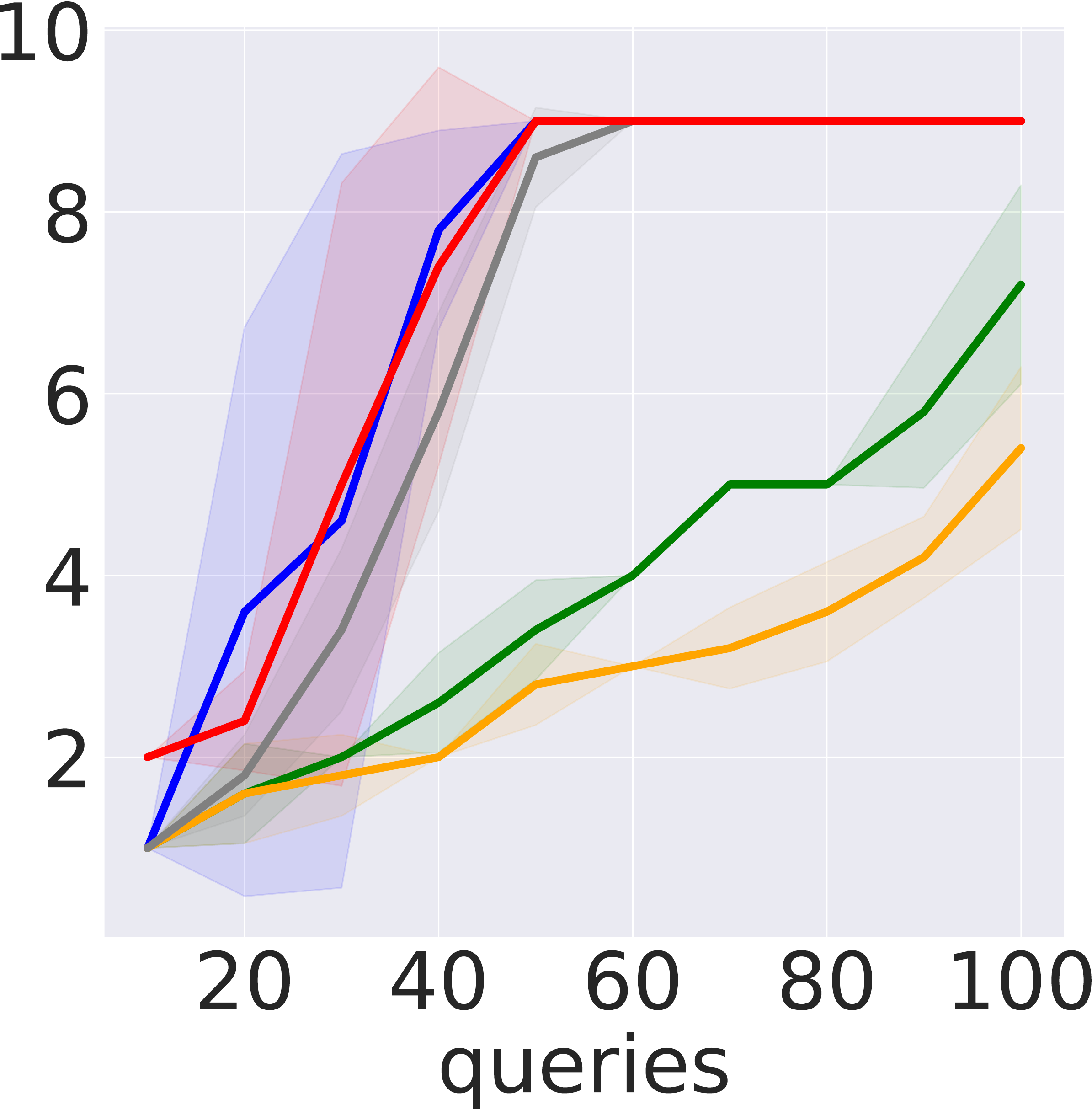}
    \caption{Glass}
  \end{subfigure}%
  \begin{subfigure}[b]{0.166\textwidth}
    \includegraphics[width=\textwidth]{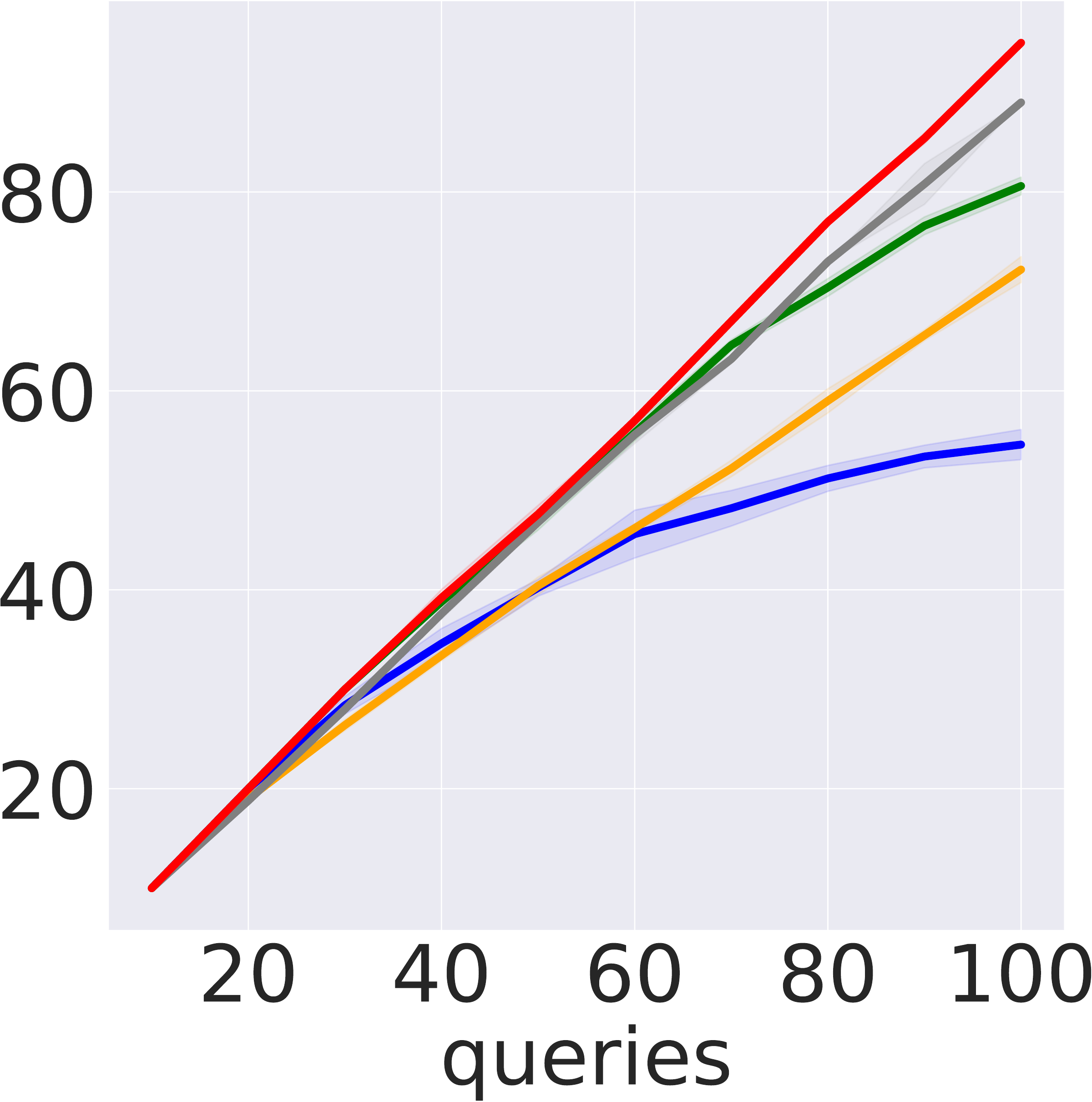}
    \caption{Ionosphere}
  \end{subfigure}%
  
  \begin{subfigure}[b]{0.166\textwidth}
    \includegraphics[width=\textwidth]{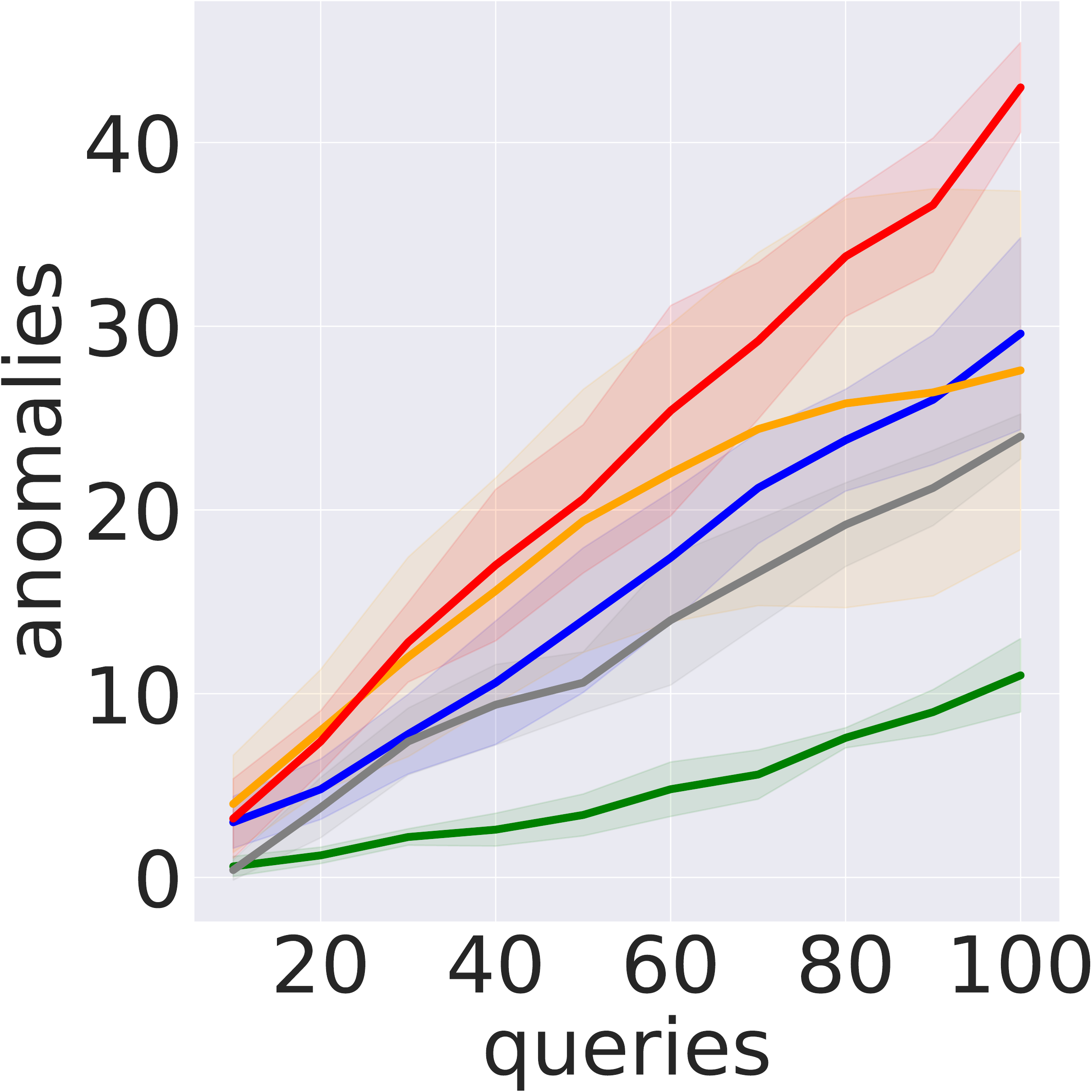}
    \caption{Letter}
  \end{subfigure}%
  \begin{subfigure}[b]{0.166\textwidth}
    \includegraphics[width=\textwidth]{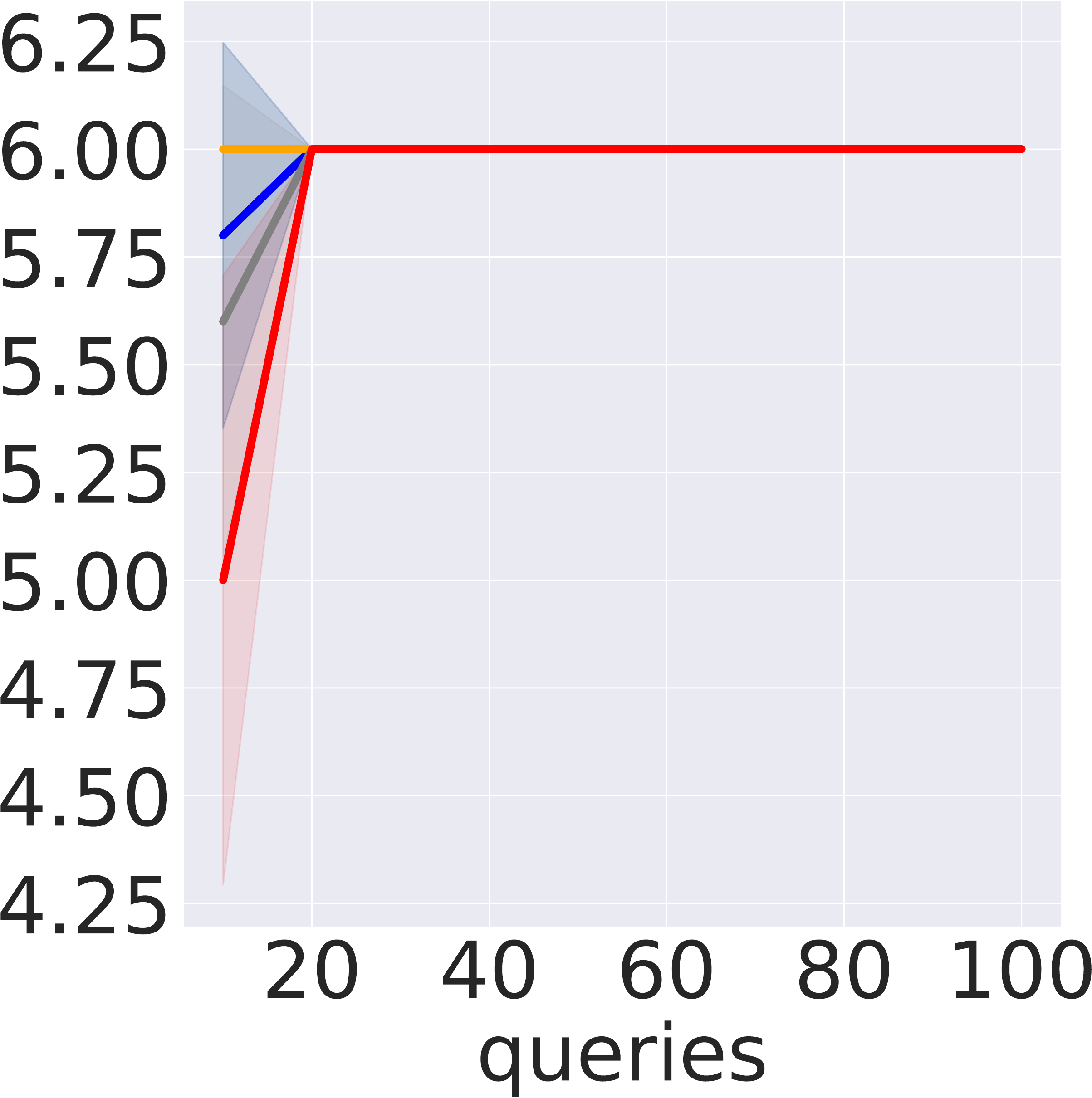}
    \caption{Lympho}
  \end{subfigure}%
  \begin{subfigure}[b]{0.166\textwidth}
    \includegraphics[width=\textwidth]{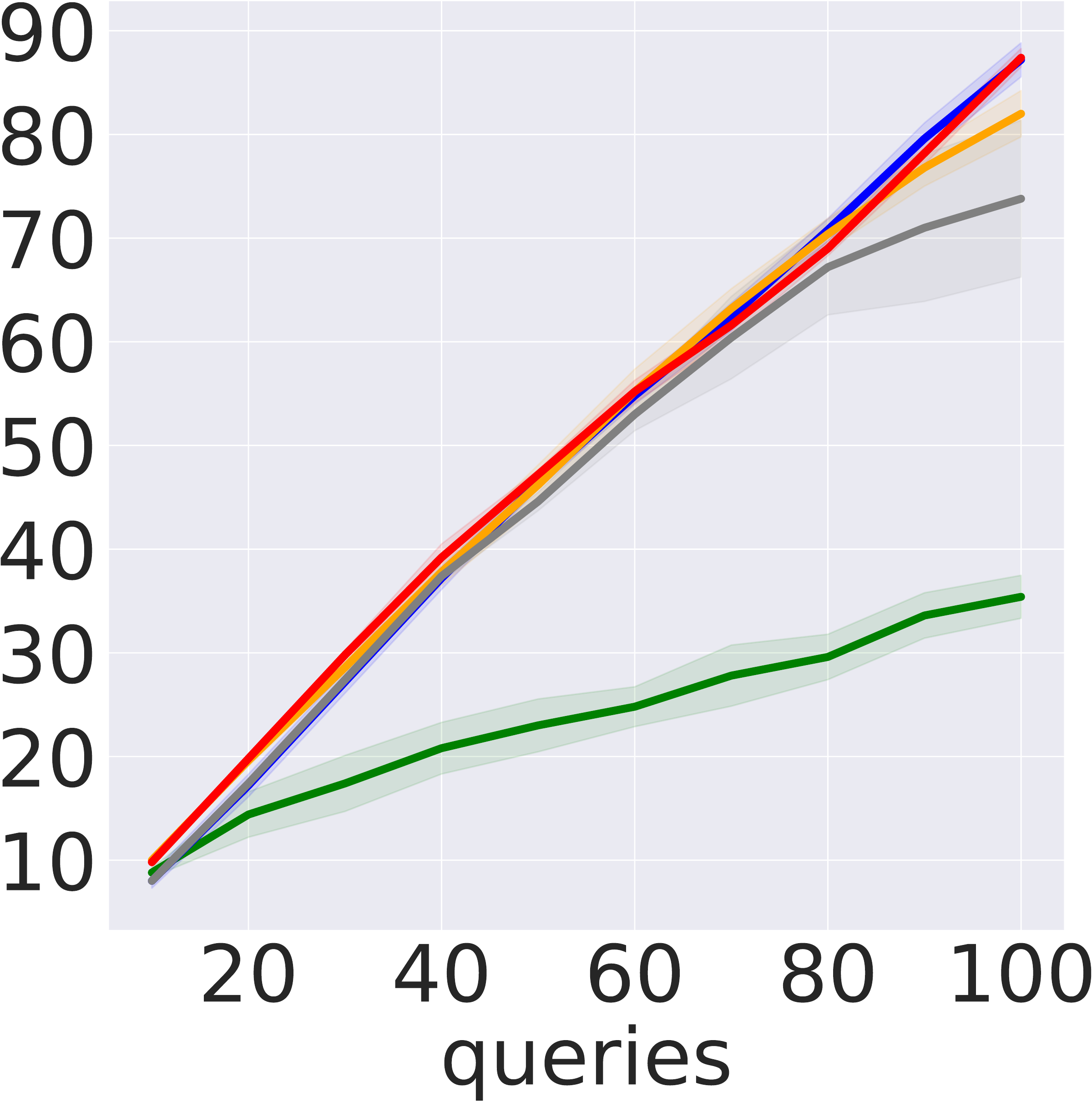}
    \caption{Mammography}
  \end{subfigure}%
  \begin{subfigure}[b]{0.166\textwidth}
    \includegraphics[width=\textwidth]{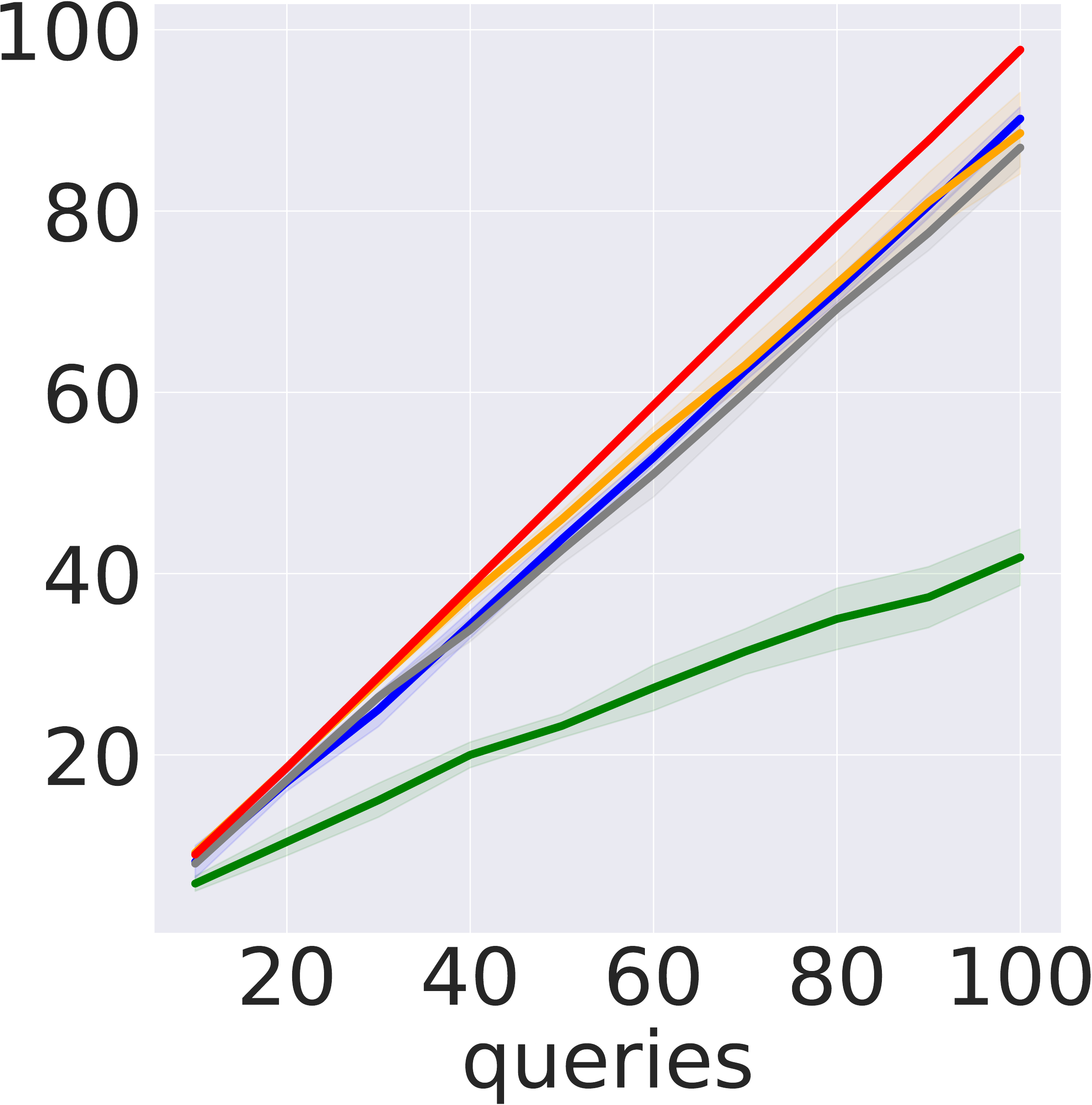}
    \caption{Mnist}
  \end{subfigure}%
  \begin{subfigure}[b]{0.166\textwidth}
    \includegraphics[width=\textwidth]{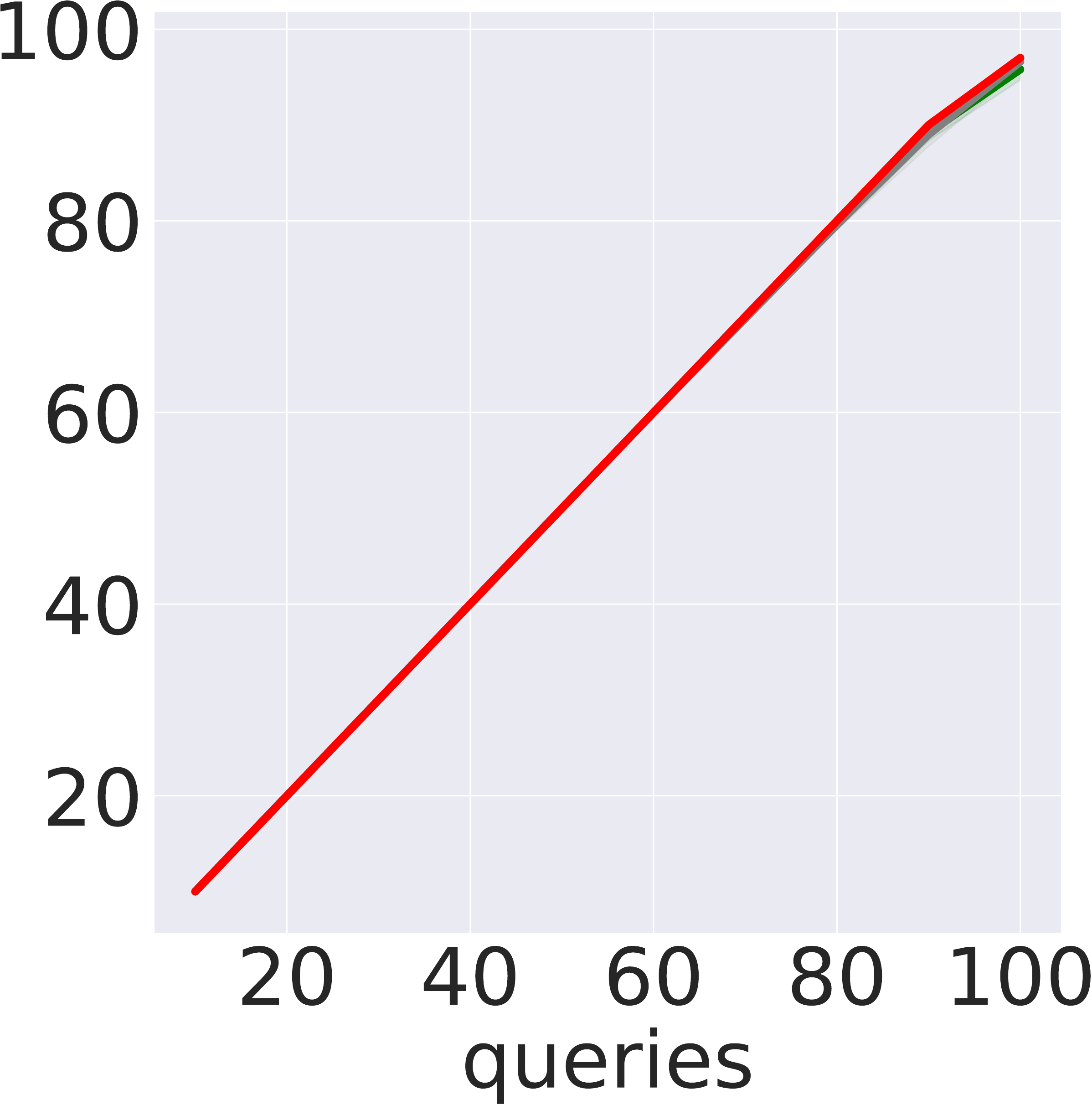}
    \caption{Musk}
  \end{subfigure}%
  \begin{subfigure}[b]{0.166\textwidth}
    \includegraphics[width=\textwidth]{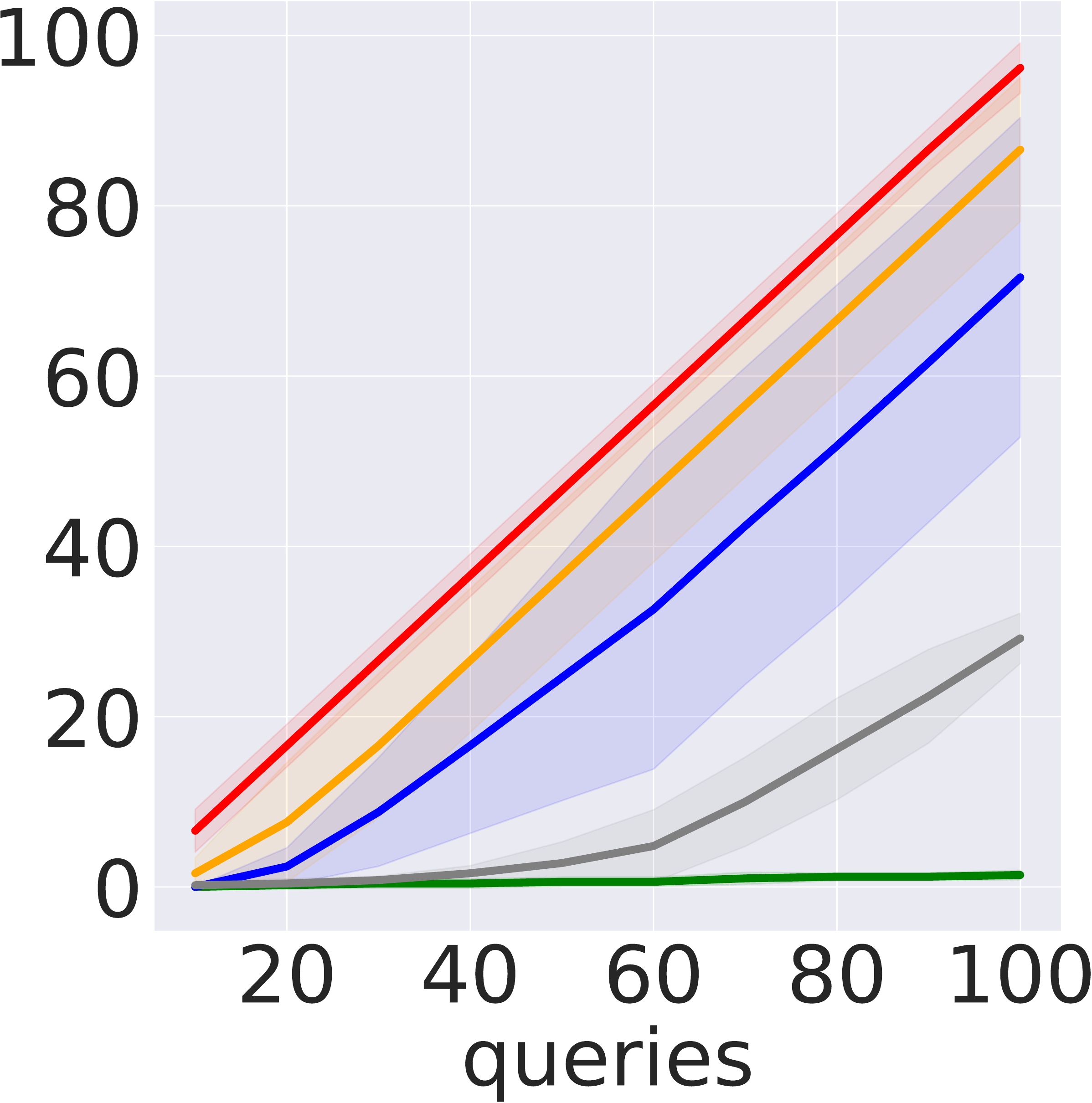}
    \caption{Optdigits}
  \end{subfigure}%
  
  \begin{subfigure}[b]{0.166\textwidth}
    \includegraphics[width=\textwidth]{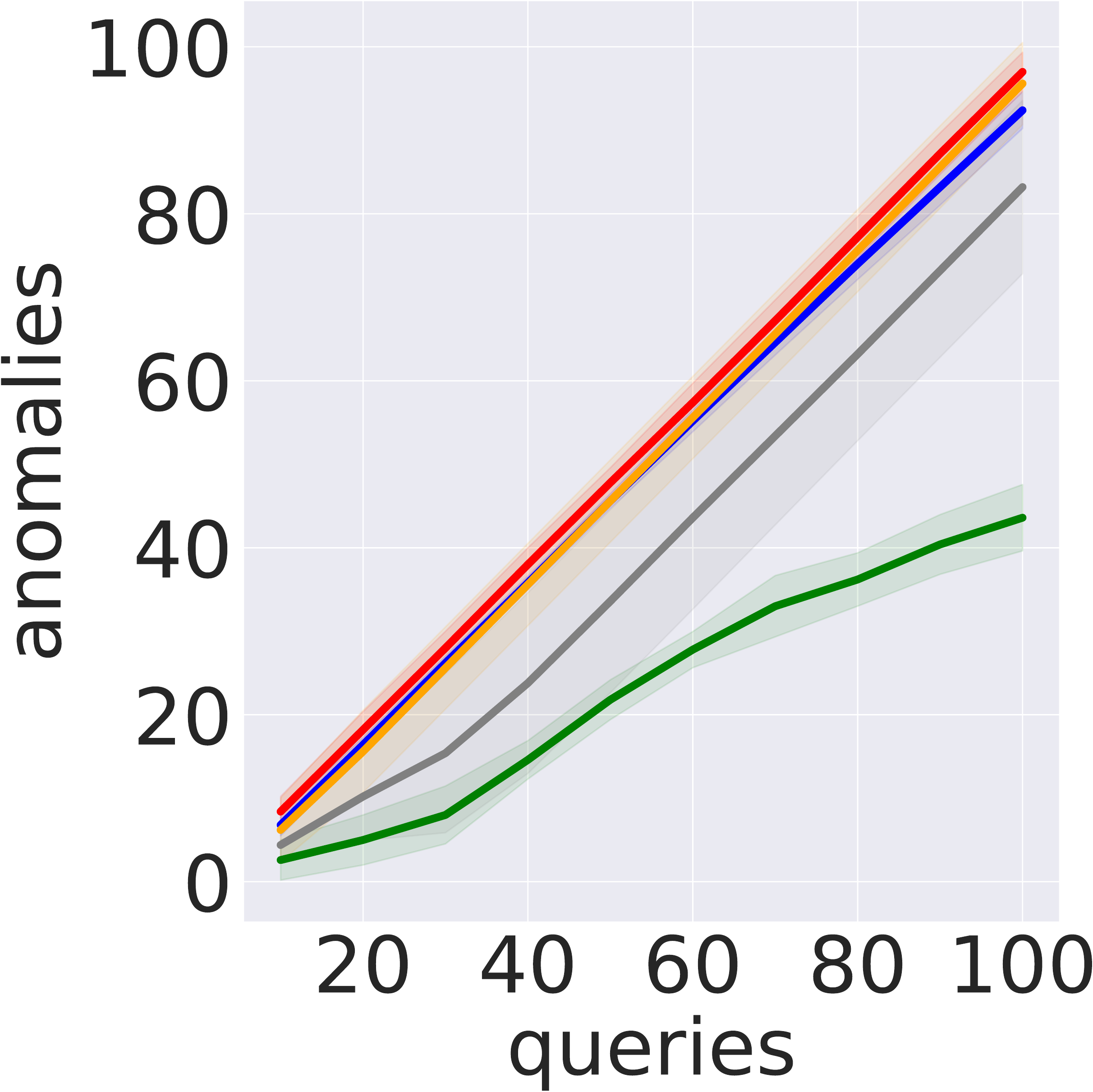}
    \caption{Pendigits}
  \end{subfigure}%
  \begin{subfigure}[b]{0.166\textwidth}
    \includegraphics[width=\textwidth]{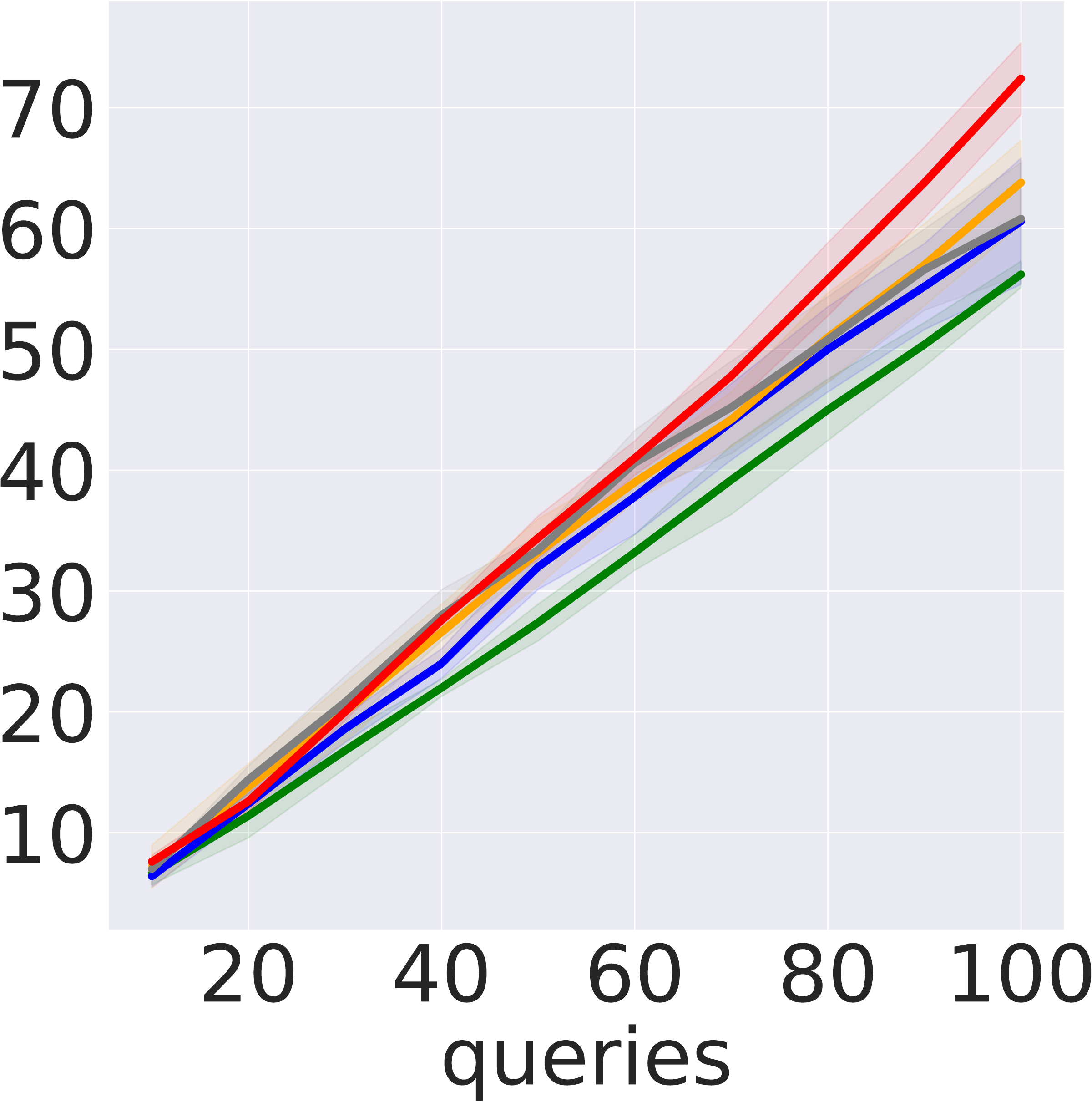}
    \caption{Pima}
  \end{subfigure}%
  \begin{subfigure}[b]{0.166\textwidth}
    \includegraphics[width=\textwidth]{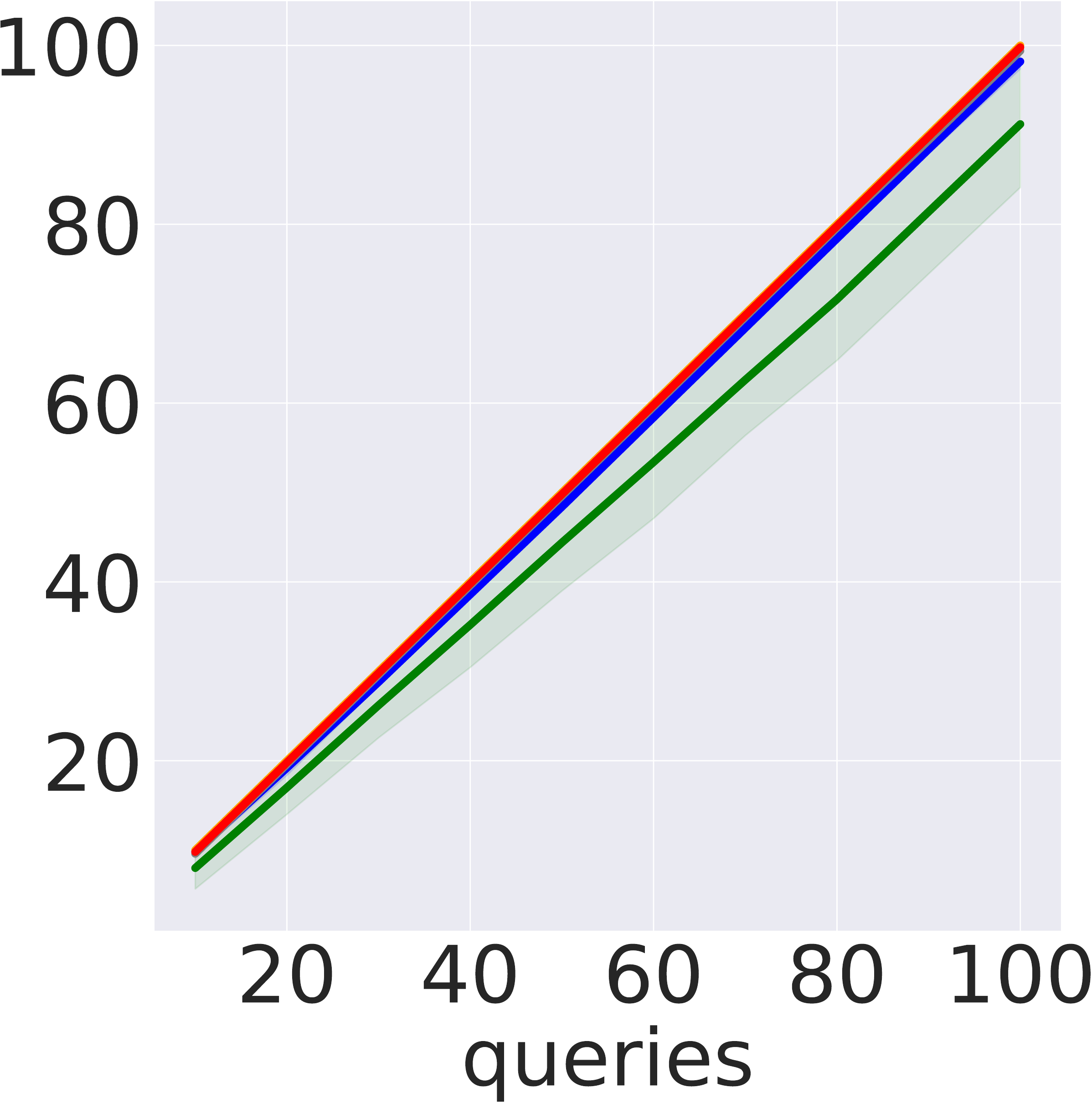}
    \caption{Satellite}
  \end{subfigure}%
  \begin{subfigure}[b]{0.166\textwidth}
    \includegraphics[width=\textwidth]{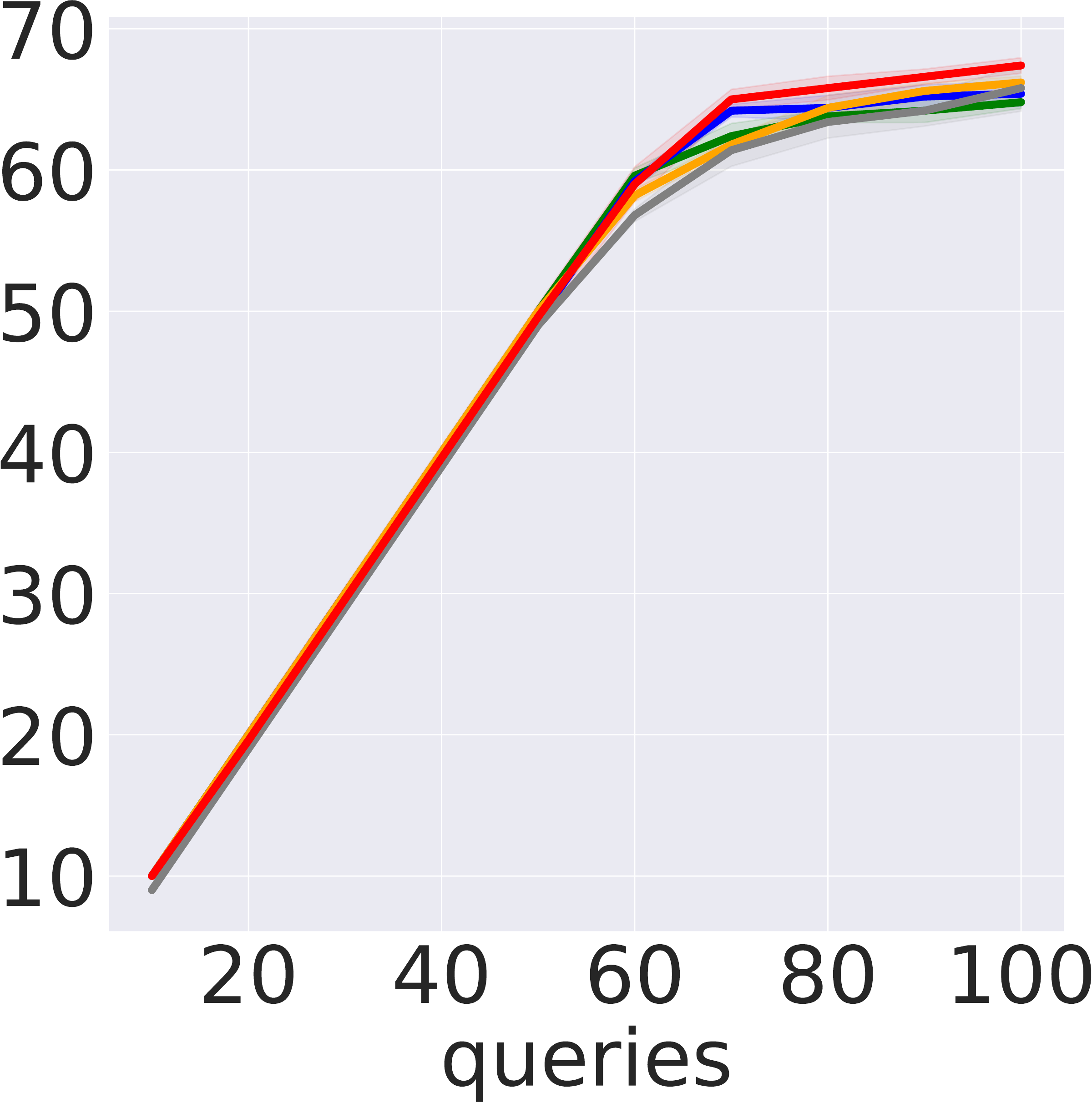}
    \caption{Satimage-2}
  \end{subfigure}%
  \begin{subfigure}[b]{0.166\textwidth}
    \includegraphics[width=\textwidth]{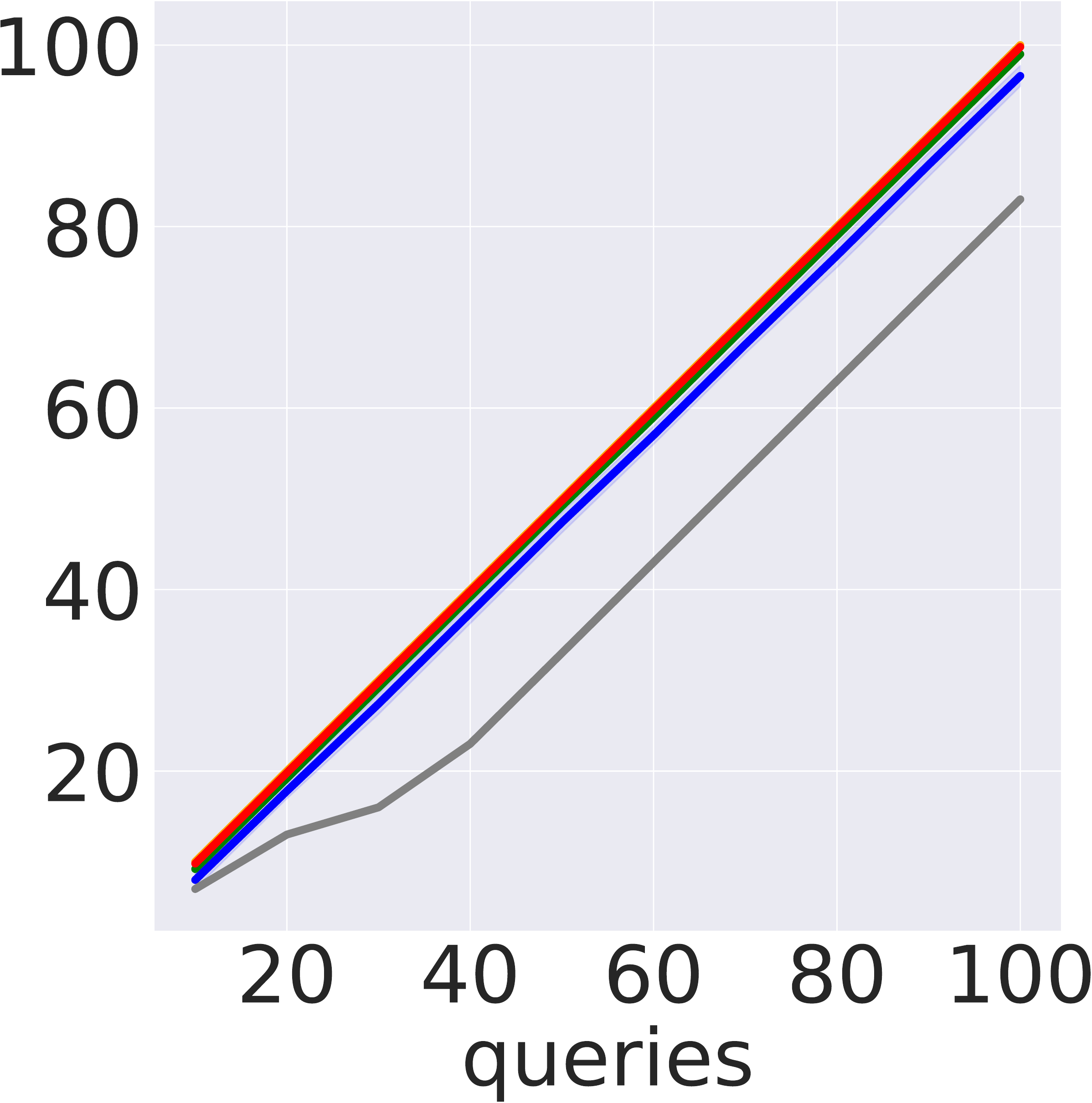}
    \caption{Shuttle}
  \end{subfigure}%
  \begin{subfigure}[b]{0.166\textwidth}
    \includegraphics[width=\textwidth]{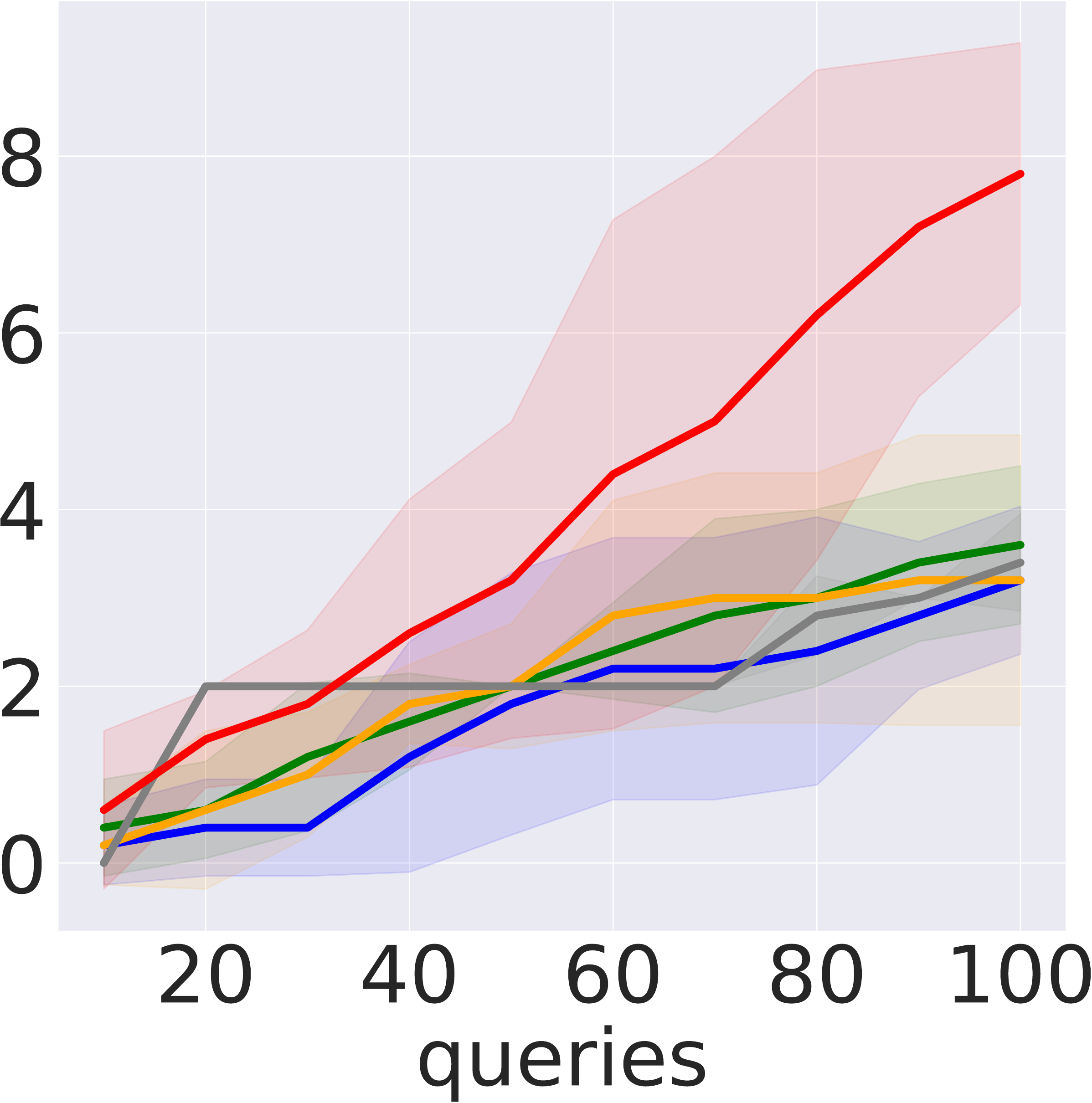}
    \caption{Speech}
  \end{subfigure}%
  
  \begin{subfigure}[b]{0.166\textwidth}
    \includegraphics[width=\textwidth]{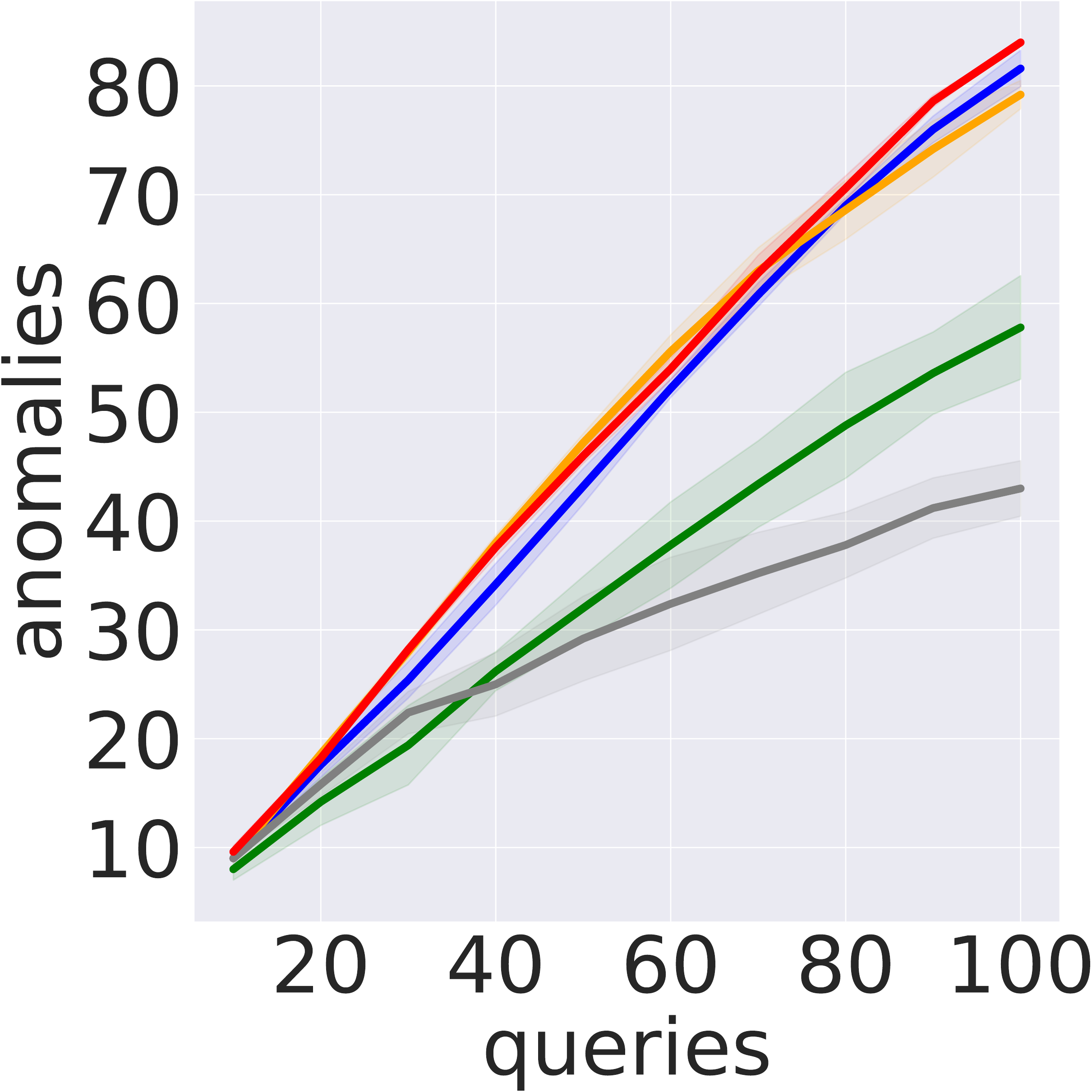}
    \caption{Thyroid}
  \end{subfigure}%
  \begin{subfigure}[b]{0.166\textwidth}
    \includegraphics[width=\textwidth]{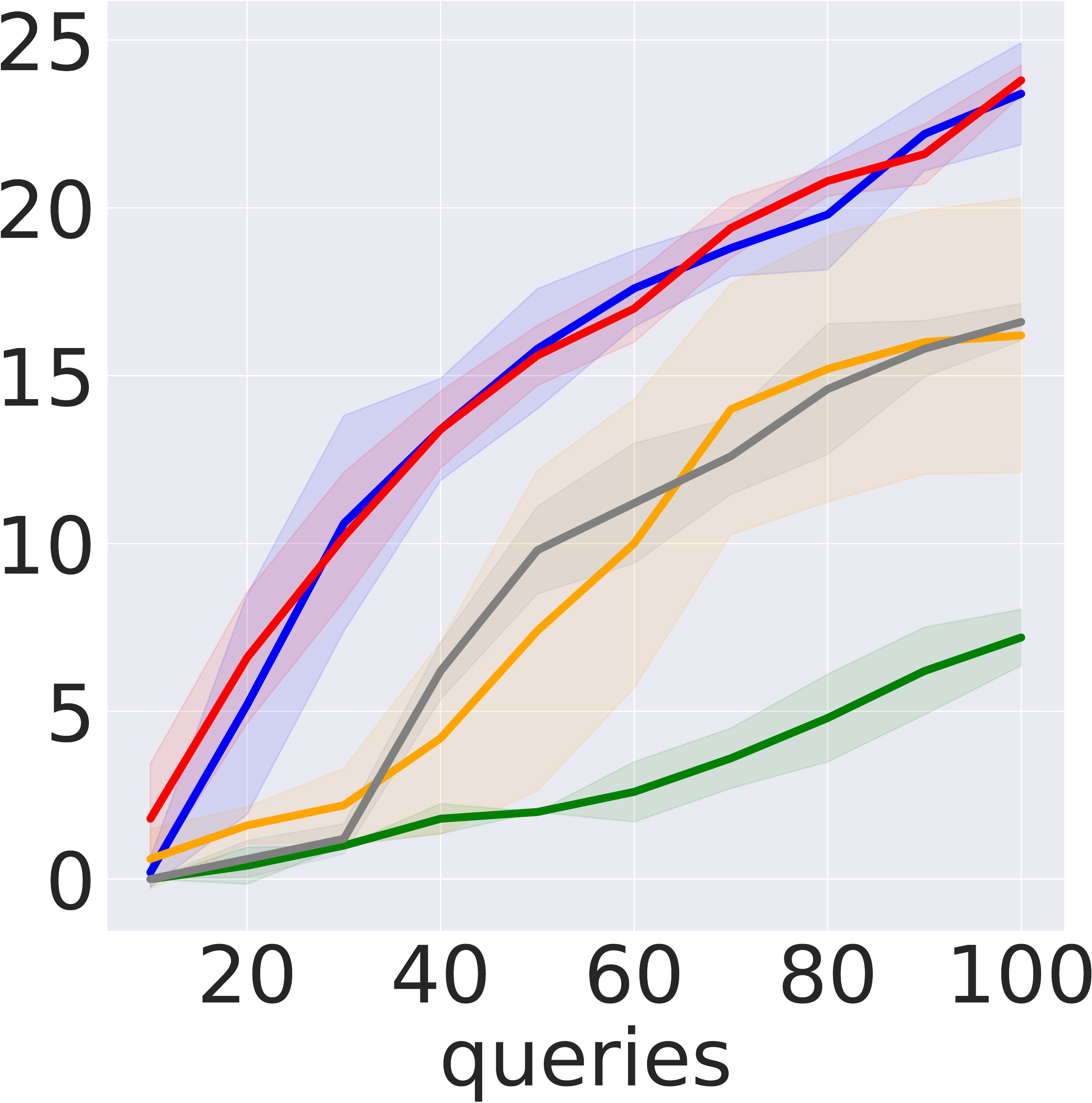}
    \caption{Vertebral}
  \end{subfigure}%
  \begin{subfigure}[b]{0.166\textwidth}
    \includegraphics[width=\textwidth]{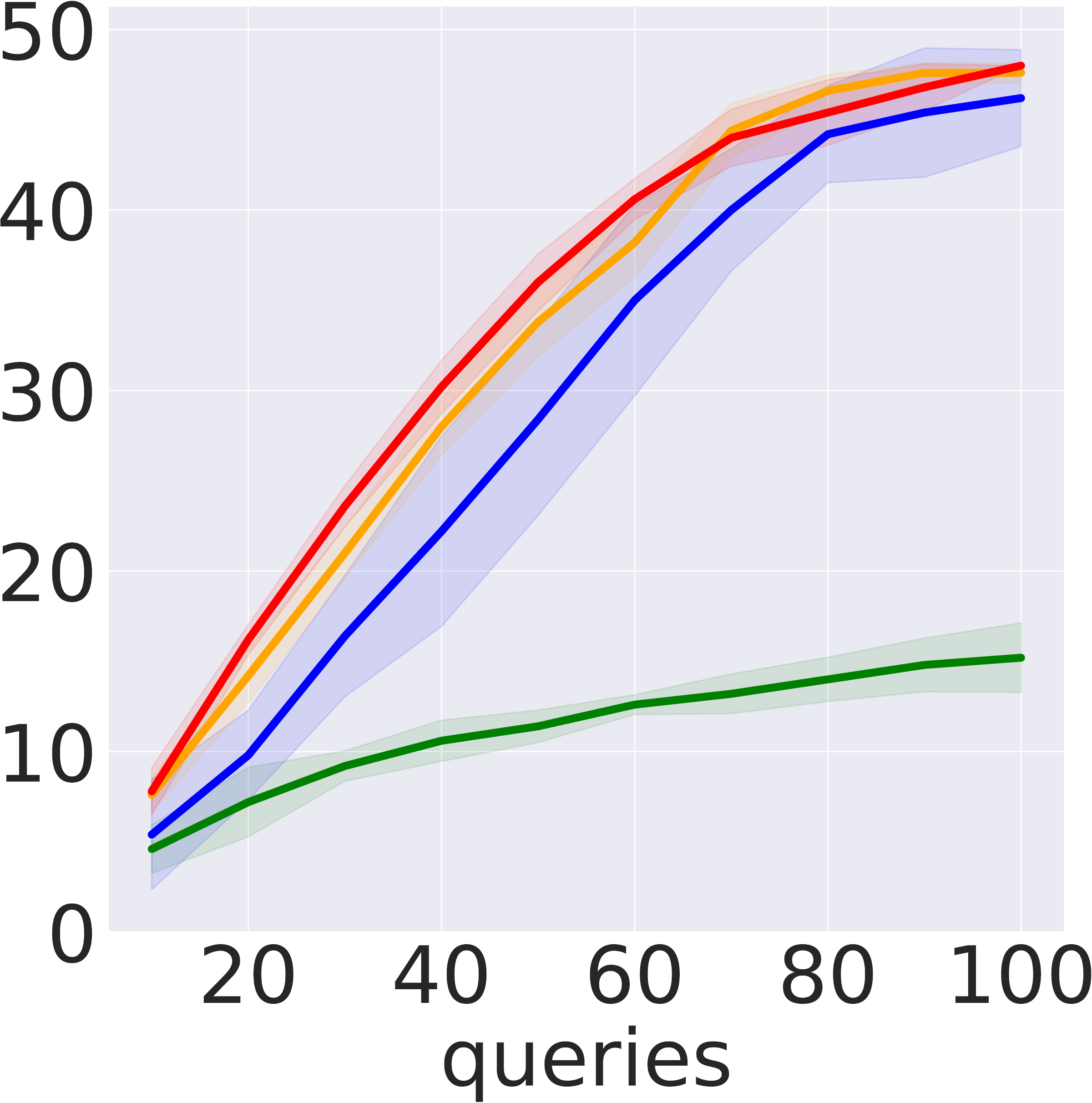}
    \caption{Vowels}
  \end{subfigure}%
  \begin{subfigure}[b]{0.166\textwidth}
    \includegraphics[width=\textwidth]{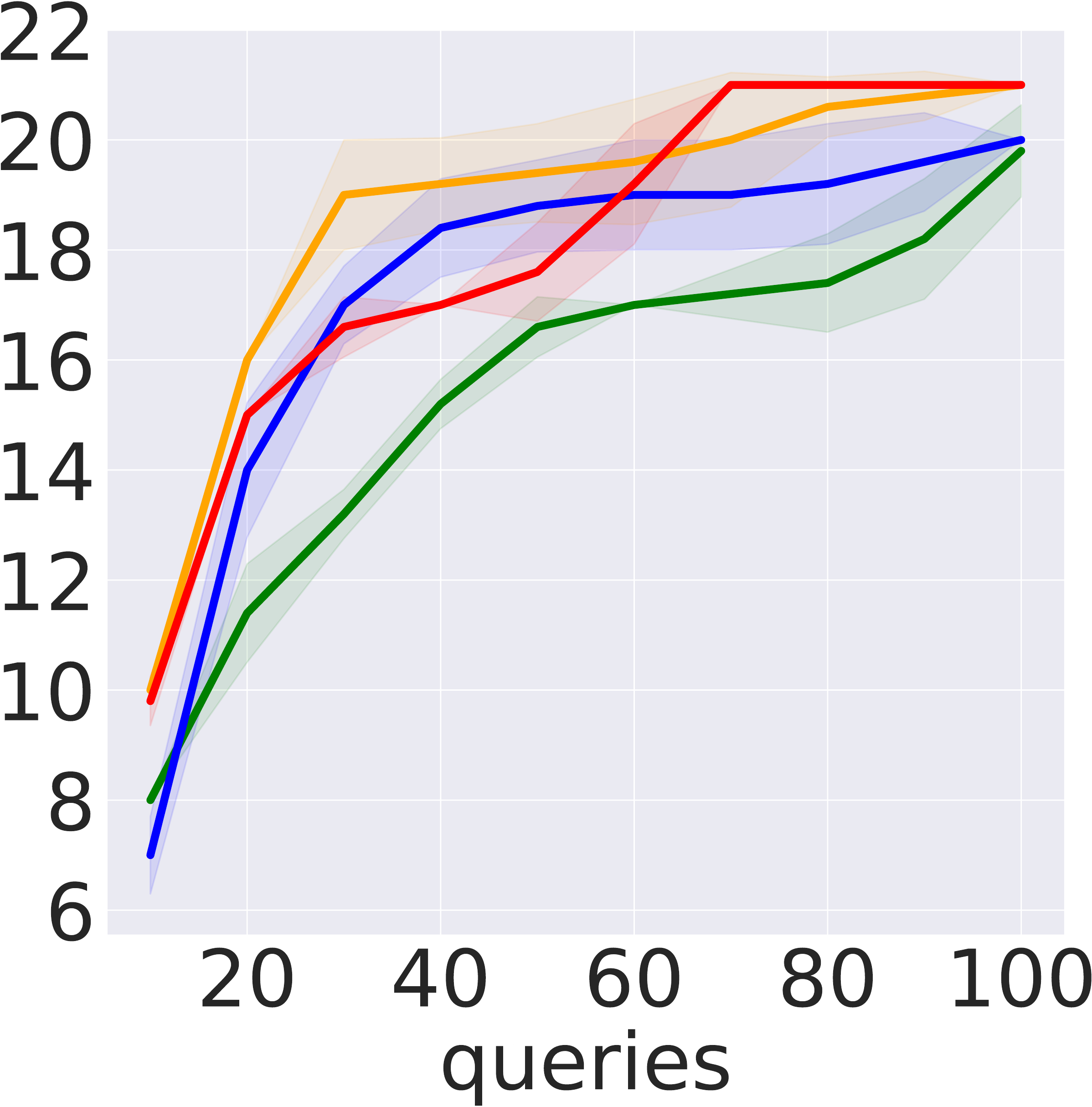}
    \caption{Wbc}
  \end{subfigure}%
  \begin{subfigure}[b]{0.166\textwidth}
    \includegraphics[width=\textwidth]{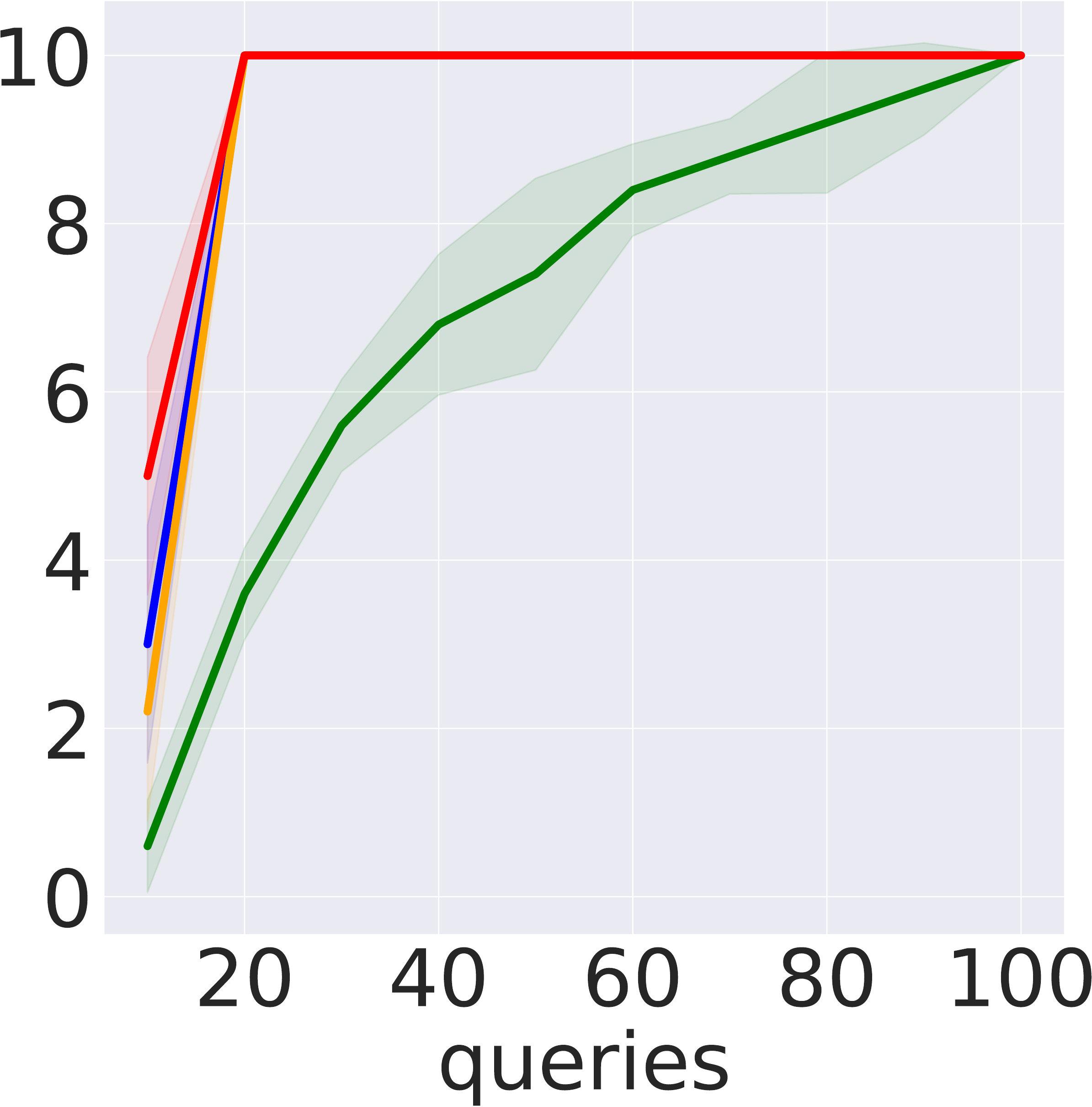}
    \caption{Wine}
  \end{subfigure}%
  \begin{subfigure}[b]{0.166\textwidth}
    \includegraphics[width=\textwidth]{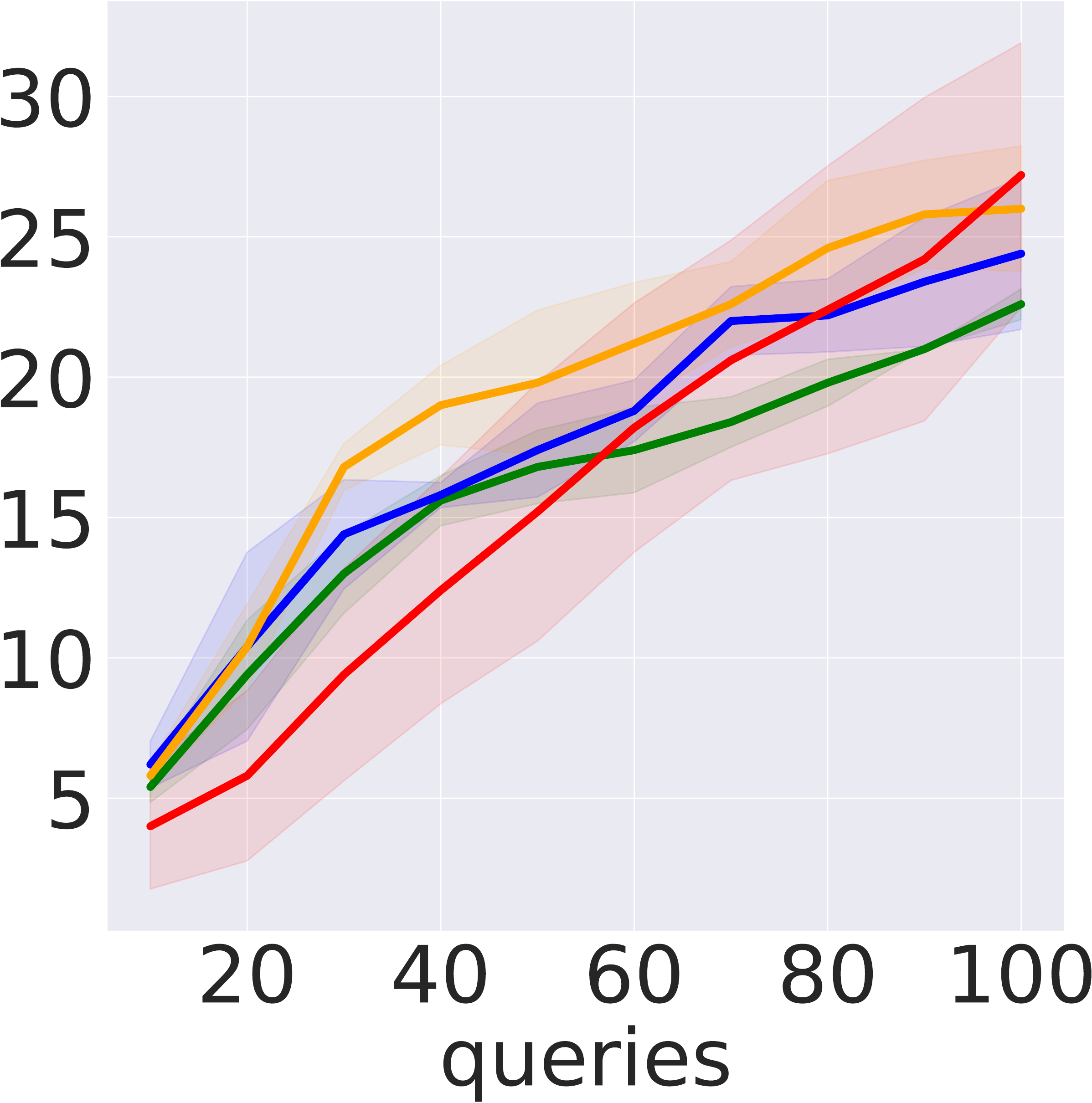}
    \caption{Yeast}
  \end{subfigure}%
    \caption{Performance comparison of Meta-AAD against the state-of-the-art alternatives and unsupervised baseline.}
  \label{fig:performance}
\end{figure*}

\textbf{Implementation details.} For training the meta-policy, we use the PPO implementation in OpenAI baselines\footnote{\url{https://github.com/hill-a/stable-baselines}}. Following the default settings, we set rollout steps $T=128$, entropy coefficient $c_2=0.01$, learning rate to be $2.5 \times 10^{-4}$, value function coefficient $c_1=0.5$, $\lambda=0.95$, clip range $\epsilon=0.2$. Recall that $\gamma$ is hyper-parameters to balance long-term and short-term rewards. We empirically set $\gamma=0.6$. We train the meta-policy with the top $12$ datasets (in alphabetical order) and apply it to the bottom $12$ datasets in Table~\ref{tab:dataset}. We do it reversely to evaluate the top $12$ datasets. The meta-policy is trained with $2 \times 10^5$ timesteps with the same hyper-parameters across all the datasets. The episode length is set to $2,000$. For the base detector of Isolation Forest, we use the implementation in sklearn\footnote{\url{https://scikit-learn.org/}} with the default hyper-parameters setting. We use the original implementations of FIF\footnote{\url{https://github.com/siddiqmd/FeedbackIsolationForest}}, AAD\footnote{\url{https://github.com/shubhomoydas/ad_examples}} and SSDO\footnote{https://github.com/Vincent-Vercruyssen/anomatools} by their authors. For FIF, we try both linear and log-likelihood losses, and report the best result. For SSDO, we find it beneficial to use Isolation Forest for the query at the beginning and then switch to SSDO when we have hit at least one anomaly. We report the results with this strategy since we observe that it outperforms randomly selecting instances at the beginning. All the experiments are run $5$ times. The average results and standard errors are reported. 

\subsection{A Case Study on the Toy Data}
\label{sec:exp2}
To study \textbf{RQ1}, we visualize the evolution of the decision of Meta-AAD on a toy data~\cite{das2017incorporating}~(see Figure~\ref{fig:toy}), which is a small dataset with $2$-dimensional features. We use the pre-trained meta-policy on the top $12$ datasets in Table~\ref{tab:dataset}. We visualize the output of action $1$ in the meta-policy, i.e., the probability of being selected for the query. Note that the probability is similar to the anomaly score, but it bases on a different objective. The top instances are expected to not only have good immediate performance, i.e., it should be very likely to be anomalous, but also benefit the performance in the long-term.

In the initial state, the meta-policy tends to choose the instances that are far away from the majority, which is similar to the behavior of unsupervised anomaly detectors. We expect that the meta-policy have learned to give more weights to detector features in the initial state when we do not have labeled samples. We can also observe that, with more queries, the decision pattern evolves. On the one hand, the probability decreases in the regions around the normal instances~(the yellow instance on the bottom left corner). On the other hand, the probability increases for the regions around anomalies~(the red triangles on the right-hand side). This behavior aligns with previous active anomaly detectors~\cite{das2017incorporating,siddiqui2018feedback}. Instead of adjusting anomaly scores, the meta-policy is optimized to maximize the discounted cumulative reward, which can better model the long-term performance compared with the previous methods.

\subsection{Performance on Benchmark Datasets}
\label{sec:exp3}
To answer \textbf{RQ2}, we compare Meta-AAD against the baselines in the $24$ real-world datasets. The anomaly discovery curves are illustrated in Figure~\ref{fig:performance}. To better understand the performance, we rank the discovered anomalies of the four algorithms under $20$, $40$, $60$, $80$ and $100$ queries, report the average rankings, and highlight the improvement of Meta-AAD over the second-best method in Table~\ref{tab:ranking}. We make the following observations.

First, all the active anomaly detectors perform significantly better than the unsupervised baseline and the semi-supervised method. Specifically, Meta-AAD, FIF and AAD can discover more anomalies using the same number of queries in the $19$ out of $24$ datasets and perform similarly in the other datasets. This is expected since labeled instances provide useful information that can help us discover more anomalies. We observe that SSDO performs slightly better than the unsupervised baseline but is far behind the active methods. A possible explanation is that SSDO optimizes a different objective and thus has sub-optimal performance in the active learning setting.

\begin{table}[]
    \centering
    \caption{Average rankings of the number of discovered anomalies under different queries across $24$ benchmarks, and the improvement of Meta-AAD over the second best state-of-the-art method. The improvement improves with more queries. Meta-AAD delivers stronger performance in long-term. $\blacktriangle$ denotes the cases where Meta-AAD is significantly better than the baseline w.r.t. the Wilcoxon signed rank test ($p<0.01$).}
    \begin{tabular}{l|lllll} \toprule
    \textbf{Method} & \textbf{20} & \textbf{40} & \textbf{60} & \textbf{80} & \textbf{100} \\ \midrule
    unsupervised~\cite{liu2008isolation}    & $4.188^{\blacktriangle}$          & $4.146^{\blacktriangle}$            & $4.167^{\blacktriangle}$          & $4.333^{\blacktriangle}$          & $4.375^{\blacktriangle}$ \\
    SSDO~\cite{vercruyssen2018semi}        & $3.312^{\blacktriangle}$          & $3.396^{\blacktriangle}$            & $3.500^{\blacktriangle}$          & $3.625^{\blacktriangle}$          & $3.438^{\blacktriangle}$ \\
    AAD~\cite{das2017incorporating}         & $3.229^{\blacktriangle}$          & $3.208^{\blacktriangle}$            & $3.271^{\blacktriangle}$          & $3.167^{\blacktriangle}$          & $3.104^{\blacktriangle}$ \\
    FIF~\cite{siddiqui2018feedback}         & $2.208$          & $2.333$            & $2.312$          & $2.396^{\blacktriangle}$          & $2.708^{\blacktriangle}$ \\
    Meta-AAD    & $\textbf{2.062}$ & $\textbf{1.917}$   & $\textbf{1.750}$ & $\textbf{1.479}$ & $\textbf{1.375}$ \\
    \midrule
    Improvement & $0.146$  & $0.416$ & $0.562$ & $0.917$ & $1.333$ \\

    \bottomrule
    \end{tabular}
    \label{tab:ranking}
\end{table}

Second, Meta-AAD consistently delivers better performance than the state-of-the-art alternatives across all the datasets. With very few exceptions, Meta-AAD improves upon the baselines. For example, Meta-AAD achieves more than $25\%$ improvement on Letter and Speech, and more than $10\%$ on Arrhythmia, Ionosphere and Pima, compared with the best alternative. In the other tasks, Meta-AAD also achieves better or similar performance. Note that Meta-AAD achieve this performance without any training or tuning on the target datasets, and thus it is easy to use in applications. The above results demonstrate the effectiveness of training a meta-policy for active anomaly detection. 

Third, Meta-Policy tends to be stronger in the long-term. In Table~\ref{tab:ranking}, we observe Meta-AAD is ranked higher and higher with more queries. Specifically, with $20$ queries, the average rank of Meta-AAD is $2.062$, which only has minor improvement over FIF. Interestingly, with $100$ queries, the average ranking of Meta-AAD becomes $1.375$. This suggests that Meta-AAD can better model long-term rewards. We speculate that deep reinforcement learning inherently models and balances short-term and long-term performance, which benefits the anomaly detector in the long-term.

\begin{figure}
  \centering
  
  \begin{subfigure}[b]{0.45\textwidth}
    \includegraphics[width=\textwidth]{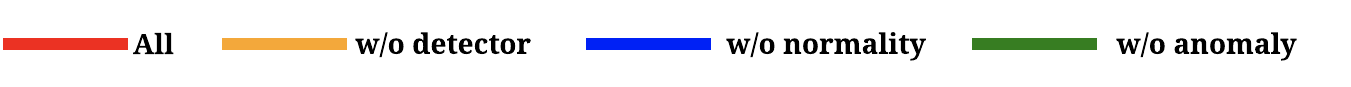}
  \end{subfigure}%

  \begin{subfigure}[b]{0.16\textwidth}
    \includegraphics[width=0.99\textwidth]{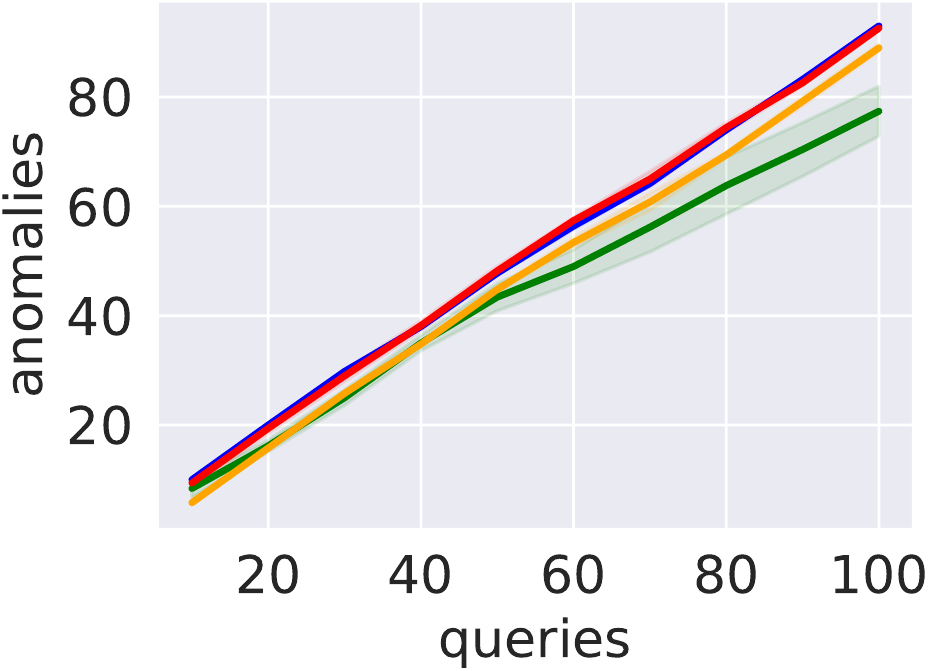}
  \end{subfigure}%
  \begin{subfigure}[b]{0.16\textwidth}
    \includegraphics[width=0.99\textwidth]{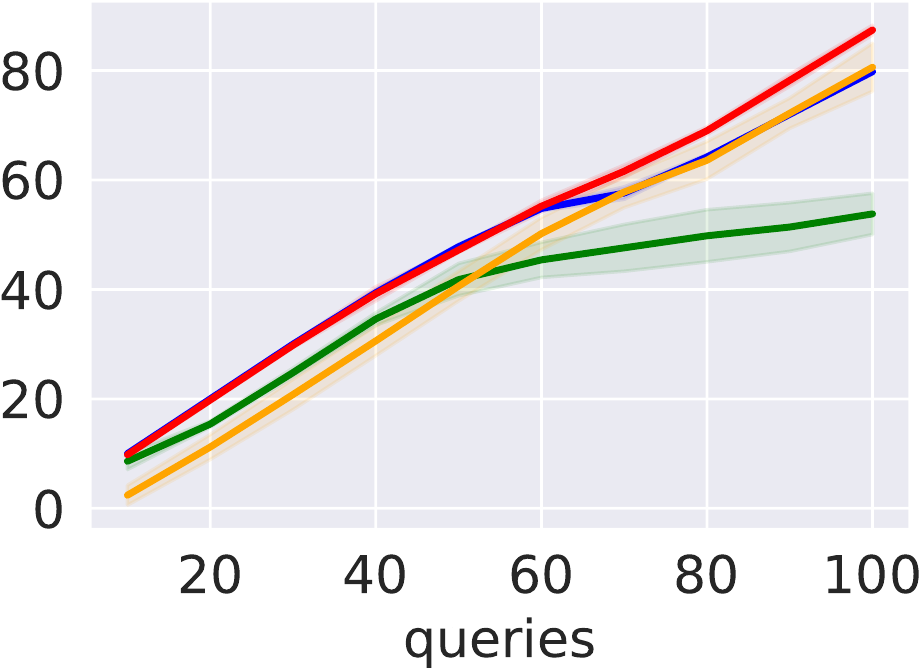}
  \end{subfigure}%
  \begin{subfigure}[b]{0.16\textwidth}
    \includegraphics[width=0.99\textwidth]{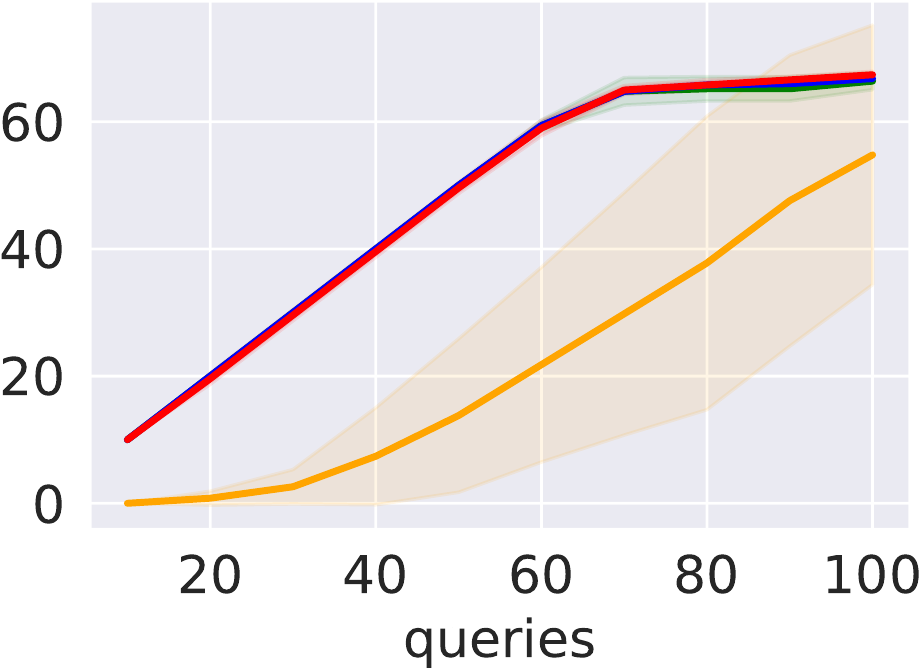}
  \end{subfigure}%
  
  \begin{subfigure}[b]{0.20\textwidth}
    \includegraphics[width=\textwidth]{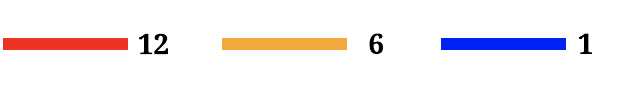}
  \end{subfigure}%
  
  \begin{subfigure}[b]{0.16\textwidth}
    \includegraphics[width=0.99\textwidth]{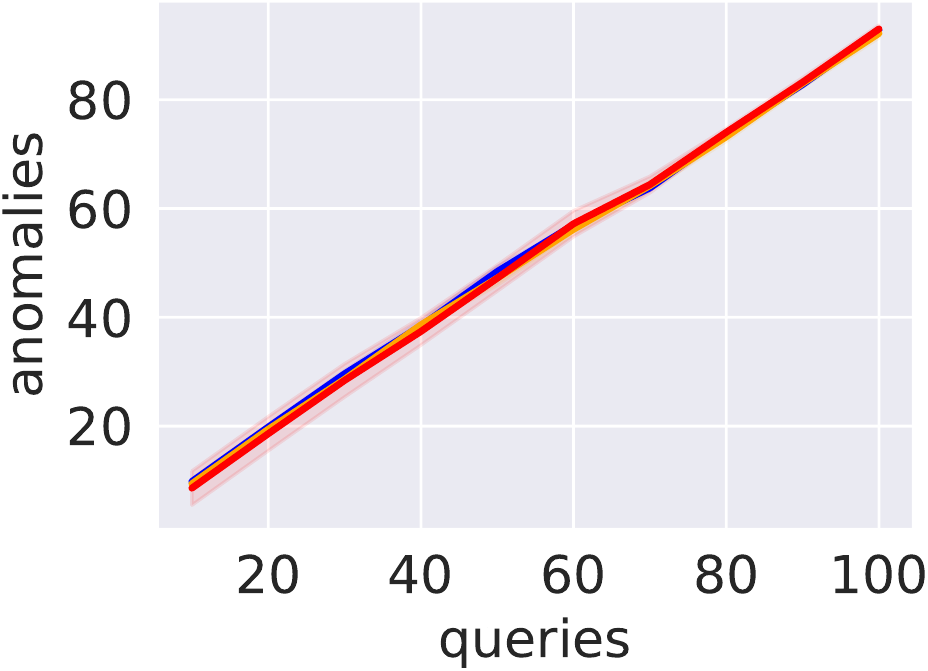}
  \end{subfigure}%
  \begin{subfigure}[b]{0.16\textwidth}
    \includegraphics[width=0.99\textwidth]{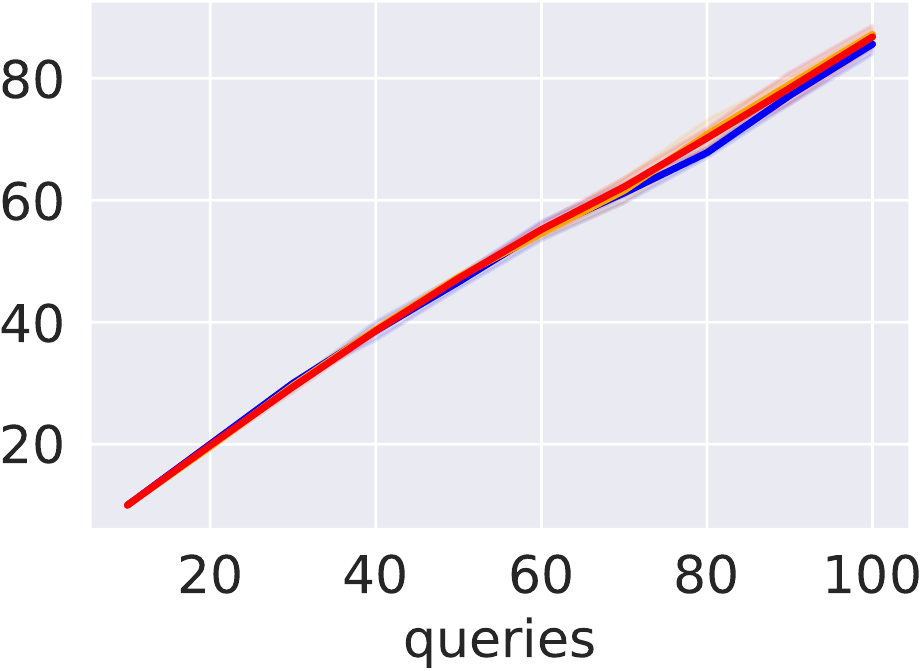}
  \end{subfigure}%
  \begin{subfigure}[b]{0.16\textwidth}
    \includegraphics[width=0.99\textwidth]{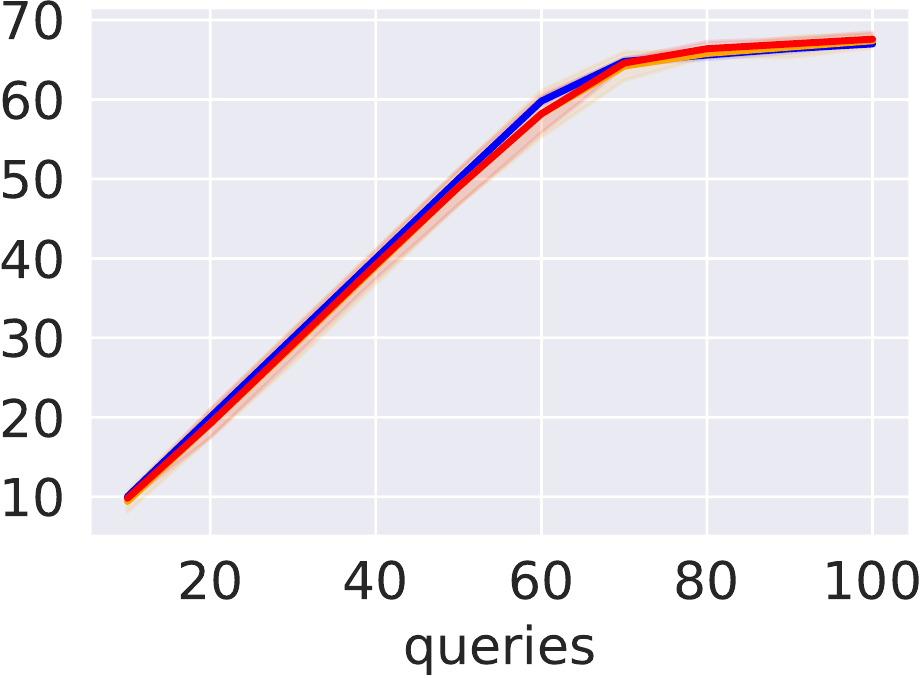}
  \end{subfigure}%
  
  \begin{subfigure}[b]{0.32\textwidth}
    \includegraphics[width=\textwidth]{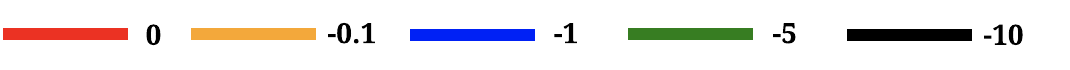}
  \end{subfigure}%
  
  \begin{subfigure}[b]{0.16\textwidth}
    \includegraphics[width=0.99\textwidth]{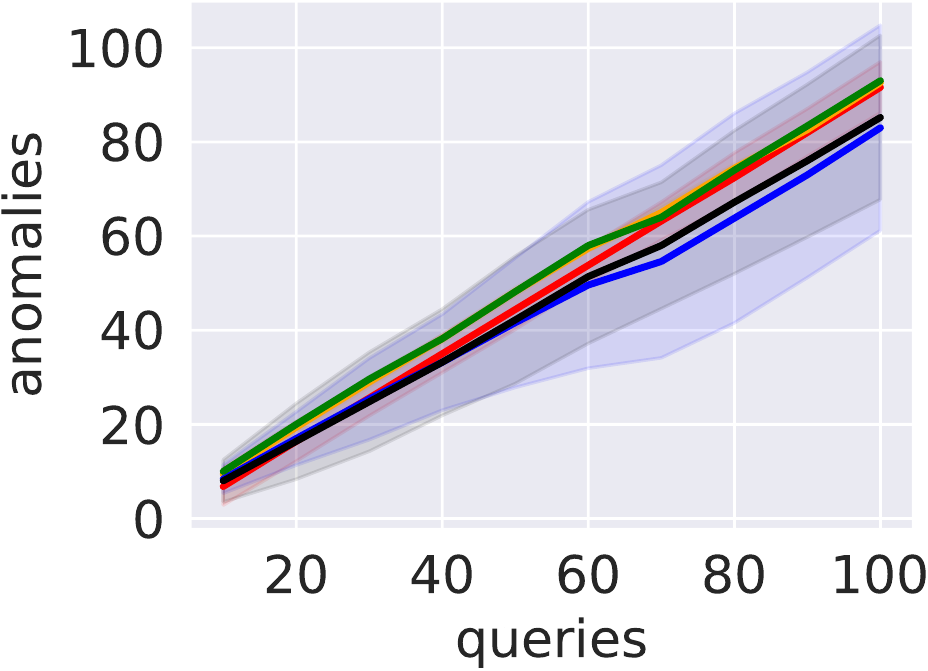}
    \caption{Annthyroid}
  \end{subfigure}%
  \begin{subfigure}[b]{0.16\textwidth}
    \includegraphics[width=0.99\textwidth]{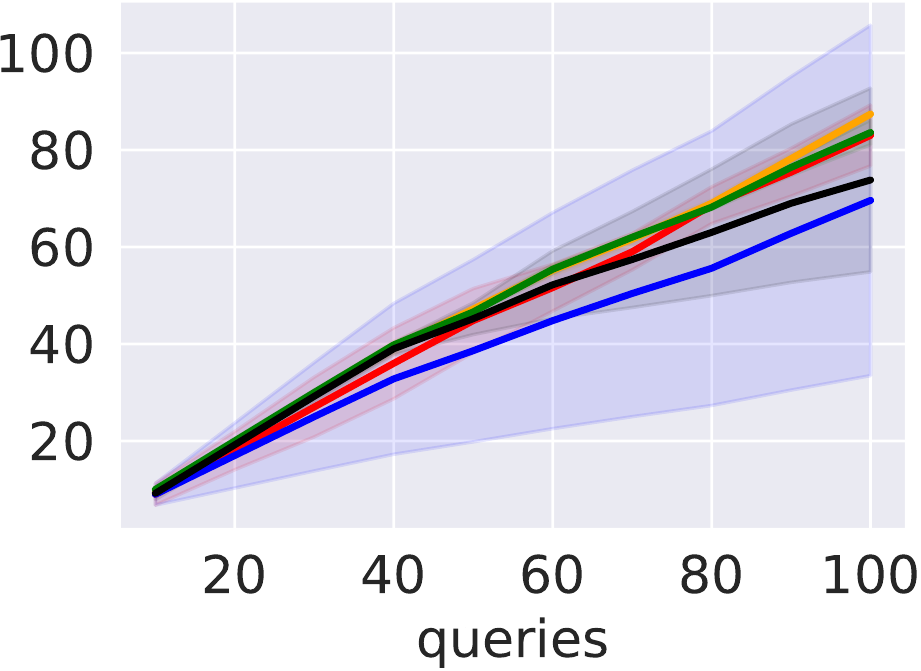}
    \caption{Mammography}
  \end{subfigure}%
  \begin{subfigure}[b]{0.16\textwidth}
    \includegraphics[width=0.99\textwidth]{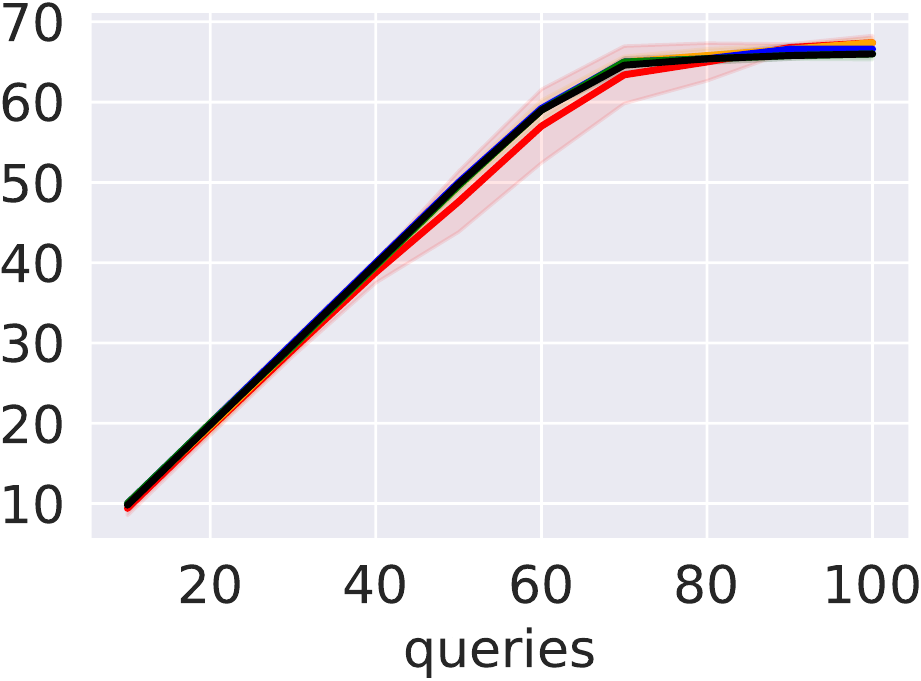}
    \caption{Satimage-2}
  \end{subfigure}%
  
  \caption{Ablation study of Meta-AAD. We show the learning curves on Annthyroid, Mammography, Satimage-2 by dropping different features (top row), using different number of training datasets (mid row), and using different negative rewards for a missed query.}
  \label{fig:ablation}
 
\end{figure}

\subsection{Ablation Studies}
\label{sec:exp4}
To better understand where the performance comes from, we answer \textbf{RQ3} with ablation studies~(see Figure~\ref{fig:ablation}). We focus on Annthyroid, Mammography, and Satimage-2.

First, we study the impact of using different features. Recall that we have three types of features, i.e., detector feature, anomaly features and normality features. We remove either of them and plot the curves in the top of Figure~\ref{fig:ablation}. We obverse that each type of feature contributes to the final performance. Using all three types of features leads to the best performance. This suggests the proposed three types of features may be complementary for training a good meta-policy. 

Second, we investigate the impact of using different number of datasets. To study whether the performance will drop if we train the meta-policy with fewer data, we report the results with $6$ and $1$ training datasets~(middle of Figure~\ref{fig:ablation}). Specifically, we randomly drop some datasets and train the meta-policy on the resulting subset. We repeat the process $20$ times and report the average performance. We observe that although the performance using more training datasets tends to be more robust, we can train a strong meta-policy even with just one dataset. This suggests that the proposed features are indeed transferable, and the proposed training strategy of the meta-policy is effective.


Third, we are interested in how the reward will impact performance. Recall that we give a positive reward of $1$ for discovered anomalies, a negative reward of $-0.1$ for selecting a normal instance, and a reward of $0$ if not querying. Here, we vary the negative rewards with other rewards fixed~(bottom of Figure~\ref{fig:ablation}). Different negative rewards will lead to different ratios between positive and negative rewards, which defines the desired behaviors of the meta-policy. We argue that the choices of the rewards should depend on the situations. For example, if examining an instance requires lots of effort, a larger negative reward is preferred. On the contrary, if we do not need many efforts to check an instance, a small negative reward could be better. As for anomaly discovery curves, we observe that too large negative rewards will worsen the performance, and a small negative reward of $-0.1$ works well across the datasets.

To summarize, we find that the default choices of work well across different datasets, delivering good performance even with few training data, which suggests that Meta-AAD could be a general framework for various scenarios.

\begin{figure}[t]
\centering

\begin{subfigure}[b]{0.20\textwidth}
  \includegraphics[width=\textwidth]{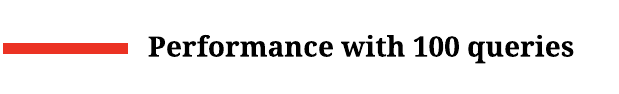}
\end{subfigure}%

\begin{subfigure}[t]{0.5\textwidth}
  \includegraphics[width=0.45\textwidth]{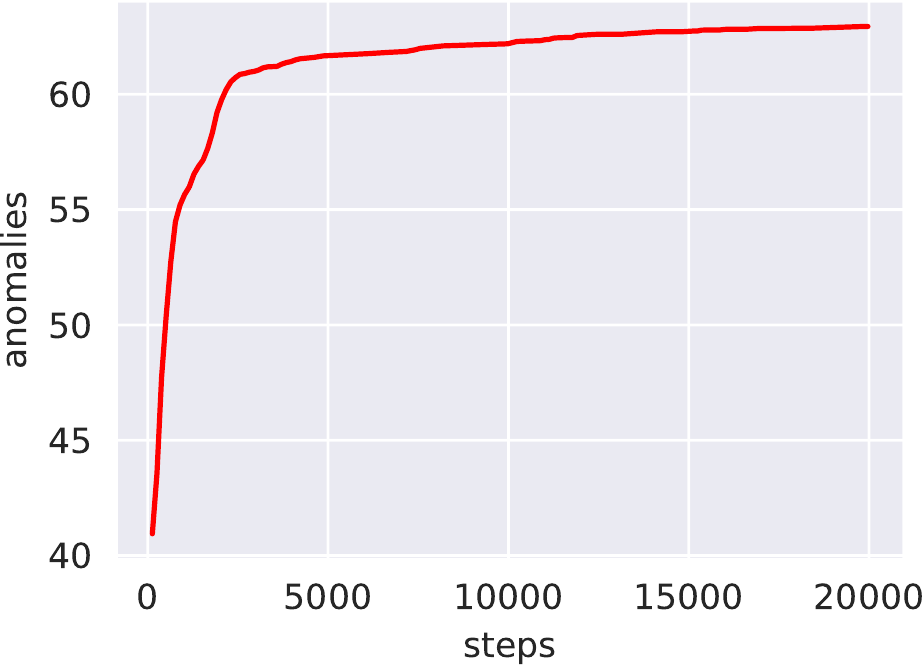}
  \includegraphics[width=0.45\textwidth]{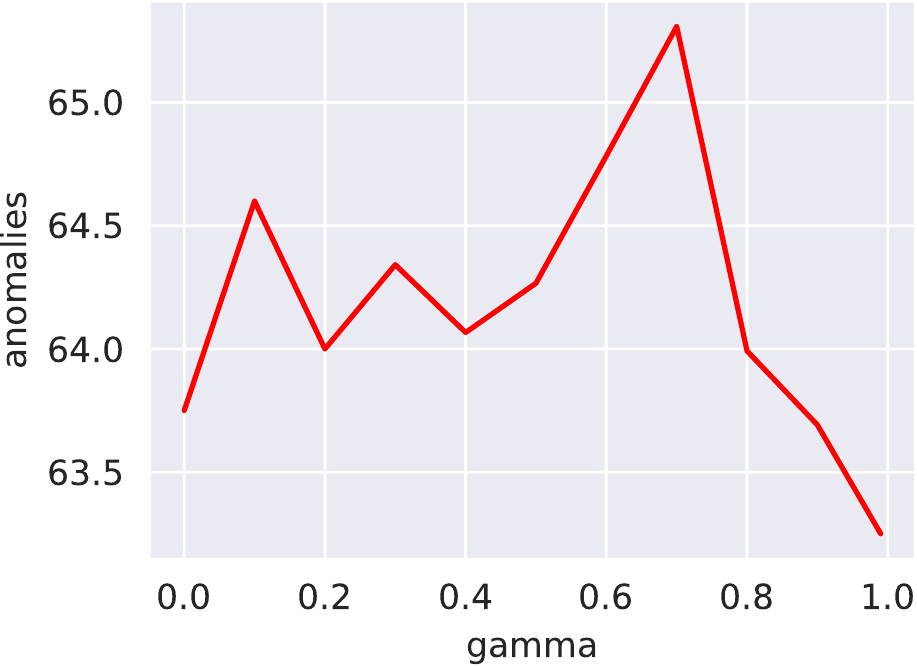}
 \end{subfigure}
  \caption{The average discovered anomalies across all the datasets given 100 queries with respect to the number of training steps (left) and different $\gamma$ values (right).}
  \label{fig:analysis}
\end{figure}

\subsection{Analysis of the Meta-Policy}
\label{sec:exp5}
We study \textbf{RQ4} by plotting the average performance with $100$ queries across the $24$ datasets with respect to the number of training steps of the meta-policy in the left-hand side of Figure~\ref{fig:analysis}. We observe that the policy converges very fast. We note that in a personal computer, it usually takes less than $30$ seconds to train $20,000$ steps with one process. Therefore, the training of the meta-policy is computationally efficient.

We investigate \textbf{RQ5} by showing the average performance with $100$ queries using different $\gamma$ in the right-hand side of Figure~\ref{fig:analysis}. Recall that $\gamma$ is a hyper-parameter to balance short-term and long-term performance. In extreme cases, $0$ suggests that we only care short-term performance, and $1$ suggests that long-term performance matters ($\gamma$ can not be larger than $1$ due to the nature of reinforcement learning algorithms). We can observe that giving too much preferences for long-term or short-term rewards will both harm the performance. We suggest that $\gamma$ should be specified based on our needs, i.e., whether we care more about long-term or short-term performance. In the conducted experiments, we set $\gamma=0.6$ across all the datasets.

\section{Related Work}
\textbf{Anomaly detection.} Anomaly detection has been extensively studied in the past decades, e.g., density-based approach~\cite{breunig2000lof}, distance-based approach~\cite{ramaswamy2000efficient,angiulli2002fast}, and ensembles~\cite{liu2008isolation,chen2017outlier,pang2018sparse}. Anomaly detection algorithms have also been developed for various types of data, such as categorical data~\cite{akoglu2012fast}, multi-dimensional data~\cite{liu2008isolation}, time-series data~\cite{gupta2013outlier} and graph data~\cite{akoglu2015graph}. Most of these algorithms are unsupervised, with strong assumptions about the anomaly patterns~\cite{zhao2019pyod}. However, these algorithms may not work well when the assumptions do not hold. On the contrary, our Meta-AAD rarely relies on the assumptions. It instead aligns anomaly patterns with human interests by leveraging human feedback

\textbf{Semi-supervised anomaly detection.} Semi-supervised learning methods~\cite{zhu2005semi,zha2019multi} have been studied in the context of anomaly detection. Semi-supervised anomaly detection assumes that a small set of labeled instances can be used to improve the performance~\cite{zhao2018xgbod}. In~\cite{tamersoy2014guilt}, a small set of anomalous instances are leveraged to re-weight the anomaly scores with belief propagation. \cite{pang2018learning} improves representation learning by using a few anomalous instances. \cite{gornitz2013toward} incorporates label information with support vector data description. AI2~\cite{veeramachaneni2016ai} ensembles unsupervised and supervised anomaly detectors. AutoML methods use a set of labeled instances to perform automated algorithm selection and neural architecture search~\cite{li2020pyodds,li2020autood}. More recently, \cite{ruff2019deep} proposes a semi-supervised anomaly detection approach for deep neural networks. However, these methods are designed for batch setting, which could be sub-optimal in the active learning.

\textbf{Active anomaly detection.} Active learning in anomaly detection is much more challenging than traditional active learning~\cite{cohn1996active,nguyen2004active} because of the imbalanced data. Instead of assuming a batch of labeled data, active anomaly detection interacts with humans and recomputes the anomaly scores based on the feedback~\cite{das2016incorporating,das2017incorporating,he2008nearest,zhou2018sparc}. These methods usually define an optimization problem based on the human feedback and re-weight the instances at each iteration. \cite{das2018active} proposes to adaptively adjust the ensemble for active anomaly detection. \cite{siddiqui2018feedback} proposes to incorporate feedback by leveraging online convex optimization to improve efficiency and simplicity. \cite{ding2019interactive} proposes to use contextual multi-armed bandit and clustering techniques to identify the anomalies in attributed networks in an interactive manner. OJRANK~\cite{lamba2019learning} re-ranks the instances in each iteration based on the top-1 feedback. While these prior methods incorporate humans in the loop, they all adopt a greedy strategy to select the top-1 anomalous instance in each iteration, which fails to model long-term performance. Whereas, our Meta-AAD builds upon deep reinforcement learning, which inherently models and optimizes long-term performance. Moreover, the previous methods require complicated optimization to re-weight the instances in each iteration. On the contrary, the trained meta-policy of meta-AAD is easy to use since it can be directly applied to different datasets without further training or tuning.

\textbf{Learning meta-policy.} Deep reinforcement learning algorithms have shown promise in various domains~\cite{mnih2015human,zha2019rlcard} The idea of meta-policy learning is to train a reinforcement learning agent to make decisions with the objective of optimizing the overall performance of the task. Some recent studies about deep reinforcement learning have demonstrated the effectiveness of the meta-policy~\cite{zha2019experience,xu2018meta,lai2020dual}. Some related studies in graph neural networks~\cite{lai2020policy} and natural language processing~\cite{duong-etal-2018-active} also show the effectiveness of meta-policy learning. In addition to the difference of objectives, these studies are limited to the same or parallel datasets. Whereas, we demonstrate that the meta-policy in Meta-AAD can be generally transferred across various datasets.

\section{Conclusions and Future Work}
In this work, we propose Meta-AAD, a framework for incorporating human feedback into anomaly detection. The meta-policy in Meta-AAD is trained with deep reinforcement learning to optimize long-term performance. We instantiate our framework with PPO and evaluate it upon $24$ benchmark datasets. The empirical results demonstrate that Meta-AAD outperforms state-of-the-art alternatives. We further conduct an extensive analysis of our framework. We find that a single configuration performs well across different datasets, and Meta-AAD can inherently balance long-term and short-term rewards, which suggests that Meta-AAD could be a general framework for active anomaly detection.

For future work, we would like to conduct more studies on how we can better extract transferable meta-features. In this work, we empirically choose $6$ features. We are interested in exploring more features to improve Meta-ADD or make the performance more robust. We would also like to try other deep reinforcement learning algorithms. Finally, we will explore the possibility of applying Meta-AAD on other tasks, such as time series, graphs, and images.

\section*{Acknowledgement}
The work is, in part, supported by NSF (IIS-1750074, CNS-1816497, IIS-1718840). The views and conclusions in this paper are those of the authors and should not be interpreted as representing any funding agencies.

\bibliographystyle{IEEEtran}
\bibliography{reference}

\begin{thebibliography}{10}
\providecommand{\url}[1]{#1}
\csname url@samestyle\endcsname
\providecommand{\newblock}{\relax}
\providecommand{\bibinfo}[2]{#2}
\providecommand{\BIBentrySTDinterwordspacing}{\spaceskip=0pt\relax}
\providecommand{\BIBentryALTinterwordstretchfactor}{4}
\providecommand{\BIBentryALTinterwordspacing}{\spaceskip=\fontdimen2\font plus
\BIBentryALTinterwordstretchfactor\fontdimen3\font minus
  \fontdimen4\font\relax}
\providecommand{\BIBforeignlanguage}[2]{{%
\expandafter\ifx\csname l@#1\endcsname\relax
\typeout{** WARNING: IEEEtran.bst: No hyphenation pattern has been}%
\typeout{** loaded for the language `#1'. Using the pattern for}%
\typeout{** the default language instead.}%
\else
\language=\csname l@#1\endcsname
\fi
#2}}
\providecommand{\BIBdecl}{\relax}
\BIBdecl

\bibitem{chandola2009anomaly}
V.~Chandola, A.~Banerjee, and V.~Kumar, ``Anomaly detection: A survey,''
  \emph{ACM computing surveys (CSUR)}, vol.~41, no.~3, pp. 1--58, 2009.

\bibitem{liu2008isolation}
F.~T. Liu, K.~M. Ting, and Z.-H. Zhou, ``Isolation forest,'' in \emph{ICDM},
  2008.

\bibitem{breunig2000lof}
M.~M. Breunig, H.-P. Kriegel, R.~T. Ng, and J.~Sander, ``Lof: identifying
  density-based local outliers,'' in \emph{SIGMOD}, 2000.

\bibitem{das2016incorporating}
S.~Das, W.-K. Wong, T.~Dietterich, A.~Fern, and A.~Emmott, ``Incorporating
  expert feedback into active anomaly discovery,'' in \emph{ICDM}, 2016.

\bibitem{das2017incorporating}
S.~Das, W.-K. Wong, A.~Fern, T.~G. Dietterich, and M.~A. Siddiqui,
  ``Incorporating feedback into tree-based anomaly detection,'' \emph{arXiv
  preprint arXiv:1708.09441}, 2017.

\bibitem{siddiqui2018feedback}
M.~A. Siddiqui, A.~Fern, T.~G. Dietterich, R.~Wright, A.~Theriault, and D.~W.
  Archer, ``Feedback-guided anomaly discovery via online optimization,'' in
  \emph{KDD}, 2018.

\bibitem{lamba2019learning}
H.~Lamba and L.~Akoglu, ``Learning on-the-job to re-rank anomalies from top-1
  feedback,'' in \emph{SDM}, 2019.

\bibitem{settles2009active}
B.~Settles, ``Active learning literature survey,'' University of
  Wisconsin-Madison Department of Computer Sciences, Tech. Rep., 2009.

\bibitem{schulman2017proximal}
J.~Schulman, F.~Wolski, P.~Dhariwal, A.~Radford, and O.~Klimov, ``Proximal
  policy optimization algorithms,'' \emph{arXiv preprint arXiv:1707.06347},
  2017.

\bibitem{mnih2015human}
V.~Mnih, K.~Kavukcuoglu, D.~Silver, A.~A. Rusu, J.~Veness, M.~G. Bellemare,
  A.~Graves, M.~Riedmiller, A.~K. Fidjeland, G.~Ostrovski \emph{et~al.},
  ``Human-level control through deep reinforcement learning,'' \emph{Nature},
  vol. 518, no. 7540, pp. 529--533, 2015.

\bibitem{schulman2015trust}
J.~Schulman, S.~Levine, P.~Abbeel, M.~Jordan, and P.~Moritz, ``Trust region
  policy optimization,'' in \emph{ICML}, 2015.

\bibitem{lillicrap2016continuous}
T.~P. Lillicrap, J.~J. Hunt, A.~Pritzel, N.~Heess, T.~Erez, Y.~Tassa,
  D.~Silver, and D.~Wierstra, ``Continuous control with deep reinforcement
  learning,'' in \emph{ICLR}, 2016.

\bibitem{dulac2015deep}
G.~Dulac-Arnold, R.~Evans, H.~van Hasselt, P.~Sunehag, T.~Lillicrap, J.~Hunt,
  T.~Mann, T.~Weber, T.~Degris, and B.~Coppin, ``Deep reinforcement learning in
  large discrete action spaces,'' \emph{arXiv preprint arXiv:1512.07679}, 2015.

\bibitem{sutton2018reinforcement}
R.~S. Sutton and A.~G. Barto, \emph{Reinforcement learning: An
  introduction}.\hskip 1em plus 0.5em minus 0.4em\relax MIT press, 2018.

\bibitem{haarnoja2018soft}
T.~Haarnoja, A.~Zhou, P.~Abbeel, and S.~Levine, ``Soft actor-critic: Off-policy
  maximum entropy deep reinforcement learning with a stochastic actor,'' in
  \emph{ICML}, 2018.

\bibitem{schulman2015high}
J.~Schulman, P.~Moritz, S.~Levine, M.~Jordan, and P.~Abbeel, ``High-dimensional
  continuous control using generalized advantage estimation,'' \emph{arXiv
  preprint arXiv:1506.02438}, 2015.

\bibitem{ding2019interactive}
K.~Ding, J.~Li, and H.~Liu, ``Interactive anomaly detection on attributed
  networks,'' in \emph{WSDM}, 2019.

\bibitem{vercruyssen2018semi}
V.~Vercruyssen, M.~Wannes, V.~Gust, M.~Koen, B.~Ruben, and D.~Jesse,
  ``Semi-supervised anomaly detection with an application to water analytics,''
  in \emph{ICDM}, 2018.

\bibitem{ramaswamy2000efficient}
S.~Ramaswamy, R.~Rastogi, and K.~Shim, ``Efficient algorithms for mining
  outliers from large data sets,'' in \emph{SIGMOD}, 2000.

\bibitem{angiulli2002fast}
F.~Angiulli and C.~Pizzuti, ``Fast outlier detection in high dimensional
  spaces,'' in \emph{ECML PKDD}, 2002.

\bibitem{chen2017outlier}
J.~Chen, S.~Sathe, C.~Aggarwal, and D.~Turaga, ``Outlier detection with
  autoencoder ensembles,'' in \emph{SDM}, 2017.

\bibitem{pang2018sparse}
G.~Pang, L.~Cao, L.~Chen, D.~Lian, and H.~Liu, ``Sparse modeling-based
  sequential ensemble learning for effective outlier detection in
  high-dimensional numeric data,'' in \emph{AAAI}, 2018.

\bibitem{akoglu2012fast}
L.~Akoglu, H.~Tong, J.~Vreeken, and C.~Faloutsos, ``Fast and reliable anomaly
  detection in categorical data,'' in \emph{CIKM}, 2012.

\bibitem{gupta2013outlier}
M.~Gupta, J.~Gao, C.~C. Aggarwal, and J.~Han, ``Outlier detection for temporal
  data: A survey,'' \emph{IEEE Transactions on Knowledge and Data Engineering},
  vol.~26, no.~9, pp. 2250--2267, 2013.

\bibitem{akoglu2015graph}
L.~Akoglu, H.~Tong, and D.~Koutra, ``Graph based anomaly detection and
  description: a survey,'' \emph{Data mining and knowledge discovery}, vol.~29,
  no.~3, pp. 626--688, 2015.

\bibitem{zhao2019pyod}
Y.~Zhao, Z.~Nasrullah, and Z.~Li, ``Pyod: A python toolbox for scalable outlier
  detection,'' \emph{arXiv preprint arXiv:1901.01588}, 2019.

\bibitem{zhu2005semi}
X.~J. Zhu, ``Semi-supervised learning literature survey,'' University of
  Wisconsin-Madison Department of Computer Sciences, Tech. Rep., 2005.

\bibitem{zha2019multi}
D.~Zha and C.~Li, ``Multi-label dataless text classification with topic
  modeling,'' \emph{Knowledge and Information Systems}, vol.~61, no.~1, pp.
  137--160, 2019.

\bibitem{zhao2018xgbod}
Y.~Zhao and M.~K. Hryniewicki, ``Xgbod: improving supervised outlier detection
  with unsupervised representation learning,'' in \emph{IJCNN}, 2018.

\bibitem{tamersoy2014guilt}
A.~Tamersoy, K.~Roundy, and D.~H. Chau, ``Guilt by association: large scale
  malware detection by mining file-relation graphs,'' in \emph{KDD}, 2014.

\bibitem{pang2018learning}
G.~Pang, L.~Cao, L.~Chen, and H.~Liu, ``Learning representations of
  ultrahigh-dimensional data for random distance-based outlier detection,'' in
  \emph{KDD}, 2018.

\bibitem{gornitz2013toward}
N.~G{\"o}rnitz, M.~Kloft, K.~Rieck, and U.~Brefeld, ``Toward supervised anomaly
  detection,'' \emph{Journal of Artificial Intelligence Research}, vol.~46, pp.
  235--262, 2013.

\bibitem{veeramachaneni2016ai}
K.~Veeramachaneni, I.~Arnaldo, V.~Korrapati, C.~Bassias, and K.~Li, ``Ai\^{} 2:
  training a big data machine to defend,'' in \emph{BigDataSecurity}, 2016.

\bibitem{li2020pyodds}
Y.~Li, D.~Zha, P.~Venugopal, N.~Zou, and X.~Hu, ``Pyodds: An end-to-end outlier
  detection system with automated machine learning,'' in \emph{WWW}, 2020.

\bibitem{li2020autood}
Y.~Li, Z.~Chen, D.~Zha, K.~Zhou, H.~Jin, H.~Chen, and X.~Hu, ``Autood:
  Automated outlier detection via curiosity-guided search and self-imitation
  learning,'' \emph{arXiv preprint arXiv:2006.11321}, 2020.

\bibitem{ruff2019deep}
L.~Ruff, R.~A. Vandermeulen, N.~G{\"o}rnitz, A.~Binder, E.~M{\"u}ller, K.-R.
  M{\"u}ller, and M.~Kloft, ``Deep semi-supervised anomaly detection,''
  \emph{arXiv preprint arXiv:1906.02694}, 2019.

\bibitem{cohn1996active}
D.~A. Cohn, Z.~Ghahramani, and M.~I. Jordan, ``Active learning with statistical
  models,'' \emph{Journal of artificial intelligence research}, vol.~4, pp.
  129--145, 1996.

\bibitem{nguyen2004active}
H.~T. Nguyen and A.~Smeulders, ``Active learning using pre-clustering,'' in
  \emph{ICML}, 2004.

\bibitem{he2008nearest}
J.~He and J.~G. Carbonell, ``Nearest-neighbor-based active learning for rare
  category detection,'' in \emph{NeurIPS}, 2008.

\bibitem{zhou2018sparc}
D.~Zhou, J.~He, H.~Yang, and W.~Fan, ``Sparc: Self-paced network representation
  for few-shot rare category characterization,'' in \emph{KDD}, 2018.

\bibitem{das2018active}
S.~Das, M.~R. Islam, N.~K. Jayakodi, and J.~R. Doppa, ``Active anomaly
  detection via ensembles,'' \emph{arXiv preprint arXiv:1809.06477}, 2018.

\bibitem{zha2019rlcard}
D.~Zha, K.-H. Lai, Y.~Cao, S.~Huang, R.~Wei, J.~Guo, and X.~Hu, ``Rlcard: A
  toolkit for reinforcement learning in card games,'' \emph{arXiv preprint
  arXiv:1910.04376}, 2019.

\bibitem{zha2019experience}
D.~Zha, K.-H. Lai, K.~Zhou, and X.~Hu, ``Experience replay optimization,'' in
  \emph{IJCAI}, 2019.

\bibitem{xu2018meta}
Z.~Xu, H.~P. van Hasselt, and D.~Silver, ``Meta-gradient reinforcement
  learning,'' in \emph{NeurIPS}, 2018.

\bibitem{lai2020dual}
K.-H. Lai, D.~Zha, Y.~Li, and X.~Hu, ``Dual policy distillation,'' in
  \emph{IJCAI}, 2020.

\bibitem{lai2020policy}
K.-H. Lai, D.~Zha, K.~Zhou, and X.~Hu, ``Policy-gnn: Aggregation optimization
  for graph neural networks,'' in \emph{KDD}, 2020.

\bibitem{duong-etal-2018-active}
L.~Duong, H.~Afshar, D.~Estival, G.~Pink, P.~Cohen, and M.~Johnson, ``Active
  learning for deep semantic parsing,'' in \emph{ACL}, 2018.

\end{thebibliography}

\end{document}